\documentclass[10pt,twocolumn,letterpaper]{article}

\usepackage[pagenumbers]{cvpr} 

\usepackage[dvipsnames]{xcolor}

\usepackage{cuted}
\usepackage{float}
\usepackage{multirow}
\usepackage{spreadtab}
\usepackage{siunitx}
\newcolumntype{H}{>{\setbox0=\hbox\bgroup}c<{\egroup}@{}}
\newcolumntype{B}{>{\bfseries}S}

\definecolor{cvprblue}{rgb}{0.21,0.49,0.74}
\usepackage[breaklinks,colorlinks,citecolor=cvprblue]{hyperref}

\def\paperID{191} 
\def\confName{3DV\xspace}
\def\confYear{2026\xspace}

\title{Evaluating Latent Generative Paradigms for High-Fidelity 3D Shape Completion from a Single Depth Image}

\author{
Matthias Humt\textsuperscript{1,2}\quad
Ulrich Hillenbrand\textsuperscript{1}\quad
Rudolph Triebel\textsuperscript{1,3}\\
\small\textsuperscript{1}German Aerospace Center\quad
\small\textsuperscript{2}TU Munich\quad
\small\textsuperscript{3}Karlsruhe Institute of Technology\\
\small\texttt{\{matthias.humt,ulrich.hillenbrand,rudolph.triebel\}@dlr.de}
}

\begin{document}
\maketitle
\begin{strip}
\centering

\includegraphics[width=0.16\linewidth]{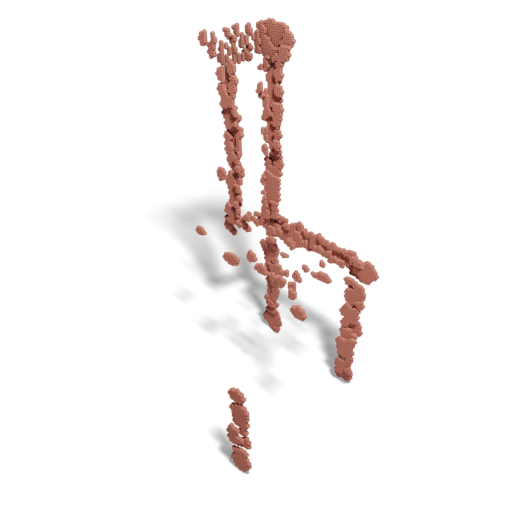}%
\includegraphics[width=0.16\linewidth]{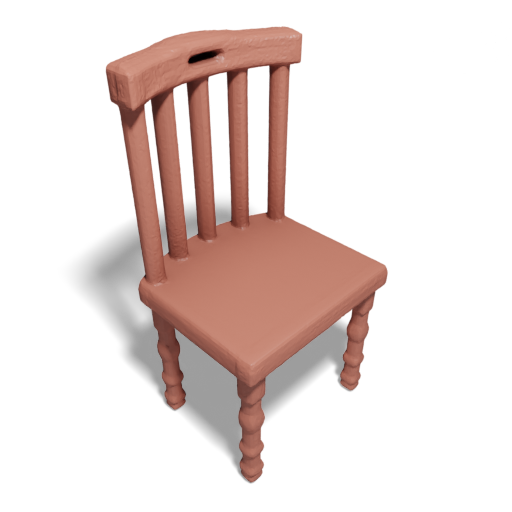}%
\includegraphics[width=0.16\linewidth]{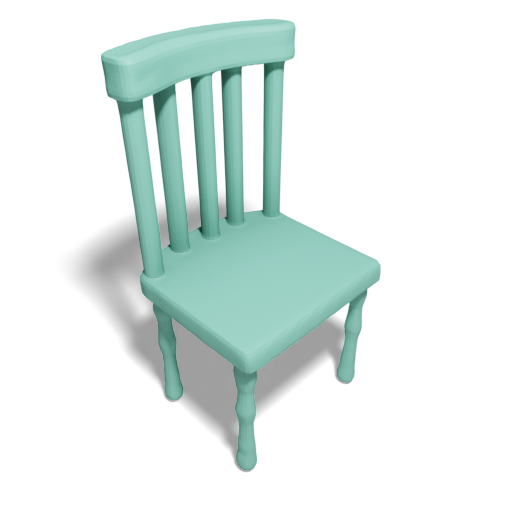}%
\includegraphics[width=0.16\linewidth]{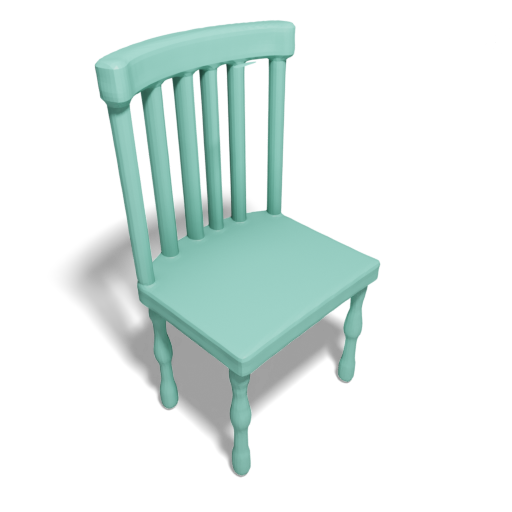}%
\includegraphics[width=0.16\linewidth]{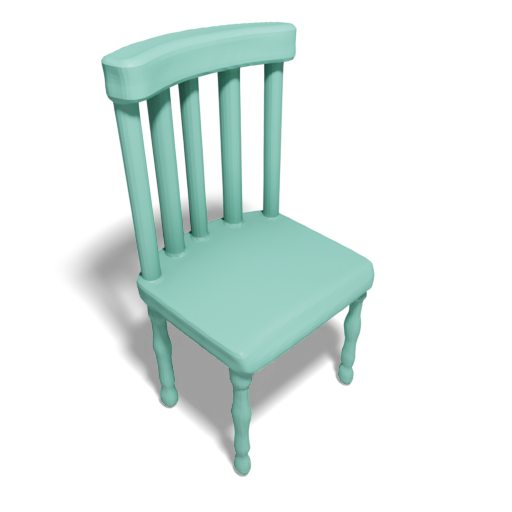}%
\includegraphics[width=0.16\linewidth]{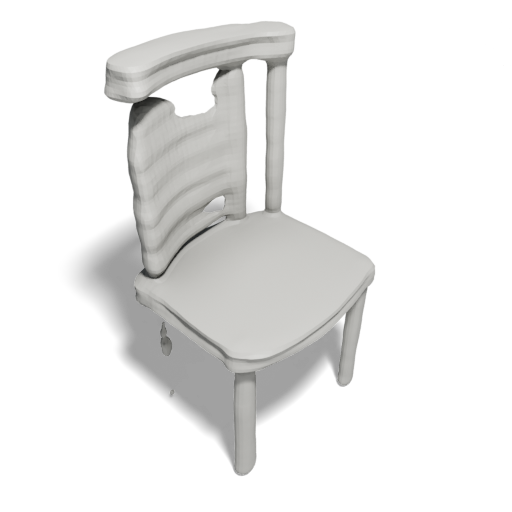}%

\includegraphics[width=0.16\linewidth]{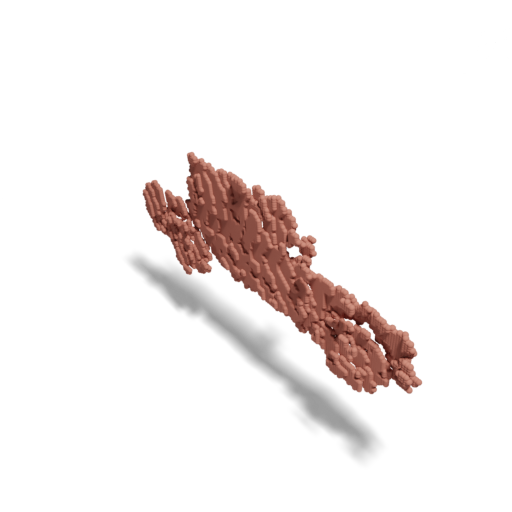}%
\includegraphics[width=0.16\linewidth]{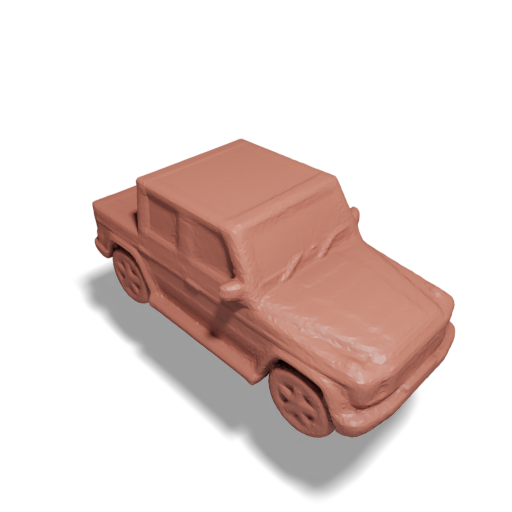}%
\includegraphics[width=0.16\linewidth]{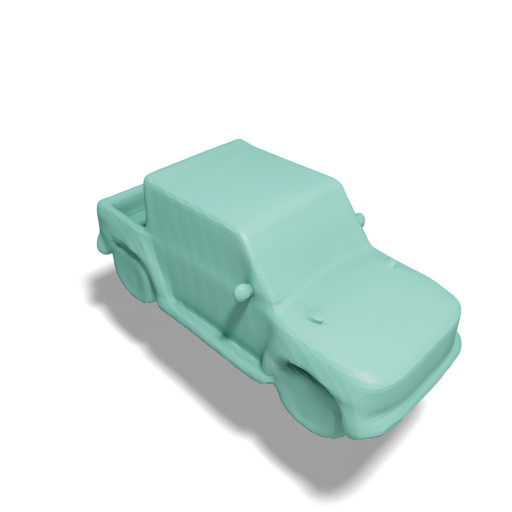}%
\includegraphics[width=0.16\linewidth]{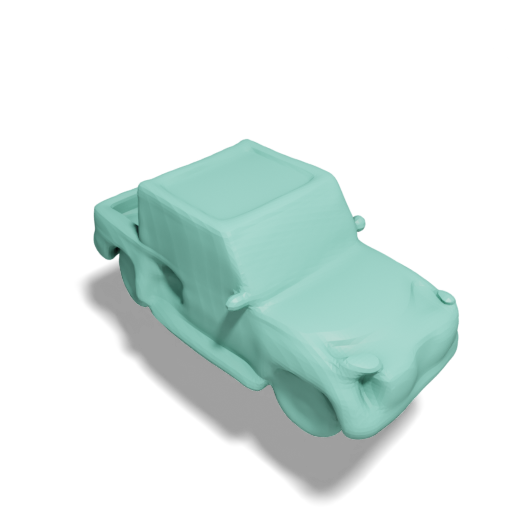}%
\includegraphics[width=0.16\linewidth]{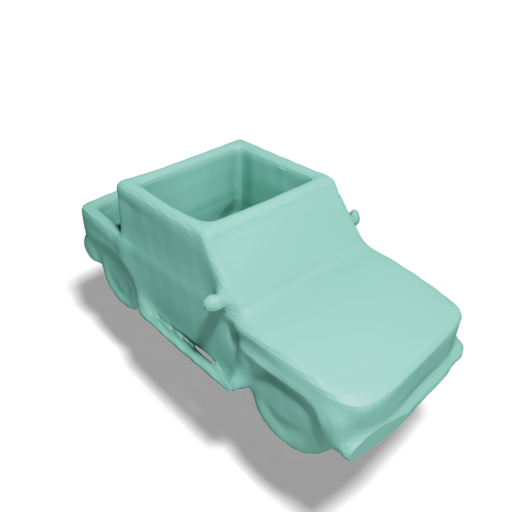}%
\includegraphics[width=0.16\linewidth]{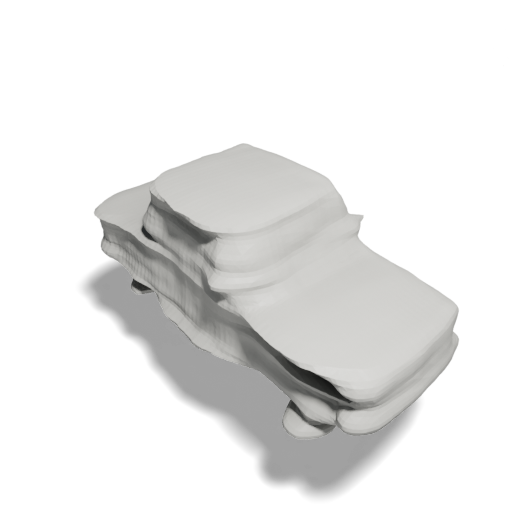}%

\includegraphics[width=0.16\linewidth]{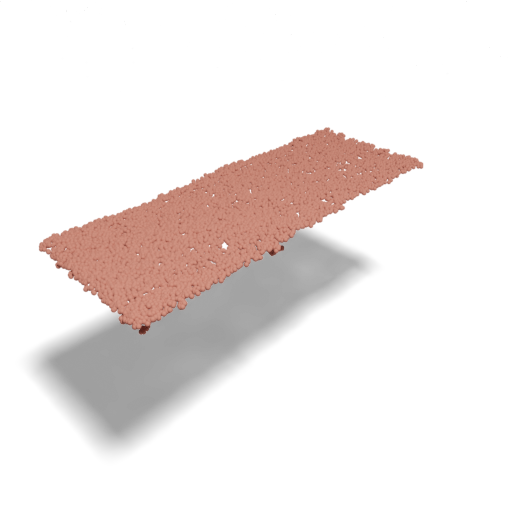}%
\includegraphics[width=0.16\linewidth]{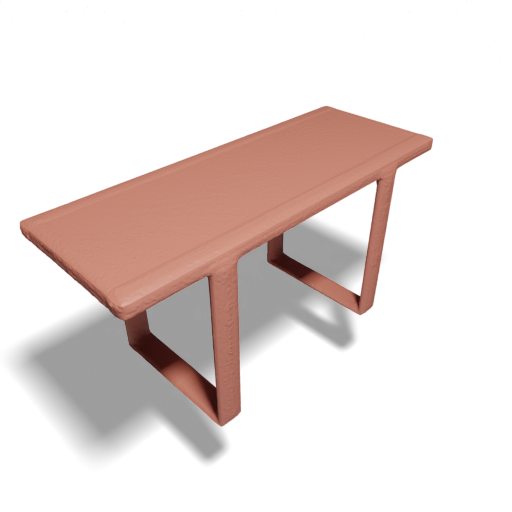}%
\includegraphics[width=0.16\linewidth]{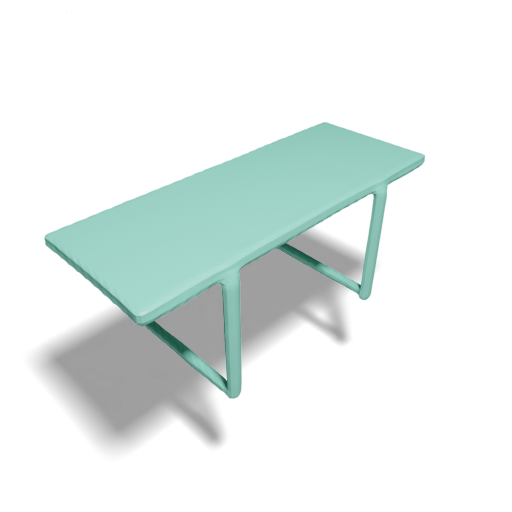}%
\includegraphics[width=0.16\linewidth]{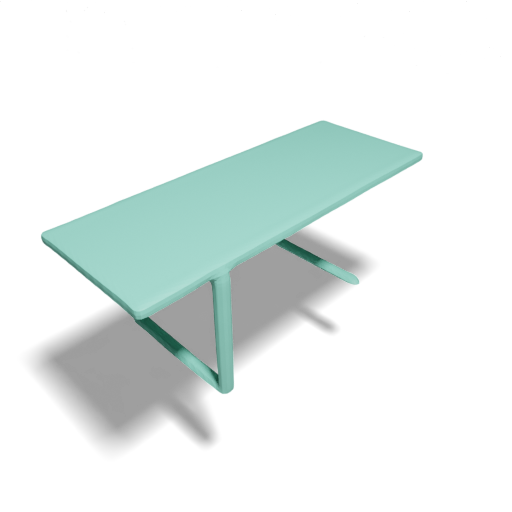}%
\includegraphics[width=0.16\linewidth]{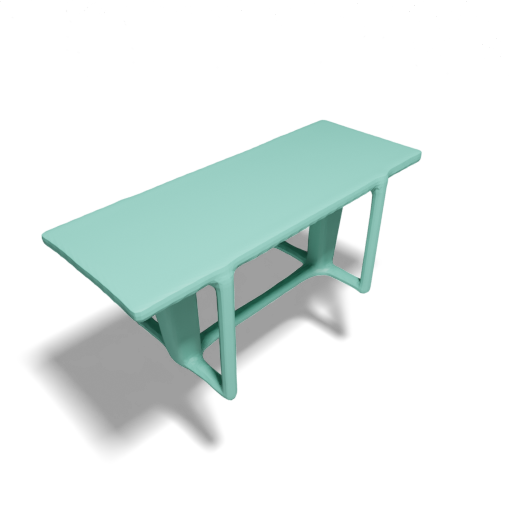}%
\includegraphics[width=0.16\linewidth]{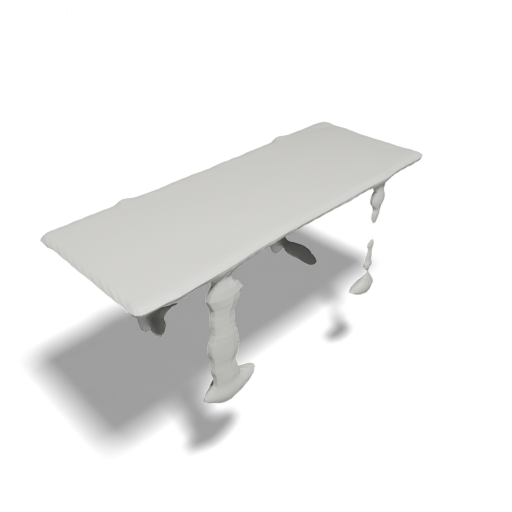}%

\captionof{figure}{Predicting complete shapes from {\color{BrickRed}\textbf{partial, noisy inputs}} (1) that closely resemble the {\color{BrickRed}\textbf{ground truth}} (2) object remains challenging when the input is highly ambiguous. We explore models that fit generative priors to latent distributions, enabling multi-modal shape completion. The generative models produce {\color{JungleGreen}\textbf{multiple plausible predictions}} (3-5) covering the range of possibilities (in descending similarity to ground truth), with some completions surpassing the quality of the {\color{darkgray}\textbf{single prediction}} (6) from discriminative models.}
\label{fig:hero}
\end{strip}

\begin{abstract}
While generative models have seen significant adoption across a wide range of data modalities, including 3D data, a consensus on which model is best suited for which task has yet to be reached. Further, conditional information such as text and images to steer the generation process are frequently employed, whereas others, like partial 3D data, have not been thoroughly evaluated. In this work, we compare two of the most promising generative models--Denoising Diffusion Probabilistic Models and Autoregressive Causal Transformers--which we adapt for the tasks of generative shape modeling and completion. We conduct a thorough quantitative evaluation and comparison of both tasks, including a baseline discriminative model and an extensive ablation study. Our results show that (1) the diffusion model with continuous latents outperforms both the discriminative model and the autoregressive approach and delivers state-of-the-art performance on multi-modal shape completion from a single, noisy depth image under realistic conditions and (2) when compared on the same discrete latent space, the autoregressive model can match or exceed diffusion performance on these tasks.
\end{abstract}

\section{Introduction}%
\label{sec:intro}

In the domain of 3D computer vision, generating complete object shapes from partial and often degraded observations is an enduring challenge, particularly for applications requiring high-fidelity, visually appealing object meshes, such as computer graphics, or accurate geometry for downstream tasks like robotics or augmented reality.

In this work, we focus on the task of single-view 3D shape completion, aiming to infer the complete 3D shape of an object from partial observations, such as a single depth image.

Many previous works have tried to tackle this problem using discriminative models~\cite{varley2017shape,yuan2018pcn,tchapmi2019topnet,yang2018dense,park2019deepsdf,chibane2020implicit,humt2023combining}, but the inherent ambiguity of the task forces these models to predict the average over all plausible completions~\cite{humt2023shape,tatarchenko2019single}, often resulting in unrealistic, low-fidelity outcomes.

Meanwhile, generative models have shown impressive results across modalities like text~\cite{radford2018improving,radford2019language,brown2020language,devlin2018bert} and audio~\cite{zeghidour2021soundstream} (1D), 2D images~\cite{ho2020denoising,karras2022elucidating} and recently also 3D data~\cite{yu2021pointr,yan2022shapeformer,zhang20223dilg,zheng2023locally,zhang20233dshape2vecset}. The latter have also been conditioned on varying modalities like text or images~\cite{zheng2023locally,zhang20233dshape2vecset} and, in some cases, limited qualitative results on partial 3D data are presented~\cite{wu20153d,yan2022shapeformer,zhang20233dshape2vecset}. Few works on generative 3D shape completion additionally provide limited quantitative evaluation~\cite{stutz2018learning,chen2019unpaired,wu2020multimodal,yu2021pointr,zhang2021unsupervised}, while none include the direct comparison to discriminative models.

The exact quantitative evaluation of generative models in general, and in the context of shape completion in particular, is still an active area of research, and a consensus on the best modeling paradigm and evaluation metrics has yet to be reached. This is notable in the variety of employed metrics and their exact definition and evaluation protocols. The situation gets aggravated by the fact that details and code for evaluation are often not provided, making it hard to reproduce and compare results.

To address this gap, we investigate two of the most promising generative models, Denoising Diffusion Probabilistic Models (DDPM)~\cite{ho2020denoising,karras2022elucidating} and Autoregressive (AR) Causal Transformers~\cite{vaswani2017attention}, on the tasks of generative shape modeling and completion. We conduct a thorough quantitative evaluation of both tasks, including a fair comparison between the two models through training on the exact same latent space and between the discriminative versus generative modeling paradigms. An extensive ablation study is also provided.
All code, weights, and data used in this work will be made publicly available upon publication.

Our main findings are: (1) Diffusion models outperform autoregressive models on both generative shape modeling and completion, which we are able to clearly attribute to the more expressive latent space of Variational Auto-Encoders~\cite{kingma2013auto} (VAE) used by the diffusion models compared to their vector-quantized variants~\cite{van2017neural} (VQ-VAE) required for latent autoregressive training. Indeed, the advantage of diffusion vanishes, and the outcome is reversed when both models are trained on the VQ-VAE latent space.
(2) Our best generative model outperforms the discriminative model in shape completion across all metrics by a large margin under correct evaluation.

We summarize the main contributions of this work as follows:

\begin{enumerate}
  \item State-of-the-art (SOTA) multi-modal shape completion from a single, noisy depth image under realistic conditions.
  \item Rigorous, \emph{quantitative} evaluation of both generative shape modeling and completion.
  \item Detailed, quantitative comparison of generative and discriminative models for shape completion.
  \item Fair, quantitative comparison of DDPMs and AR Causal Transformers for shape modeling and completion.
  \item A runtime-optimized reference implementation of the evaluation protocol, including a large number of commonly used metrics.
\end{enumerate}
\section{Related Work}%
\label{sec:related}

\textbf{Discriminative shape modeling.} Early works, enabled by the advent of large 3D object datasets~\cite{chang2015shapenet}, predicted shapes using 3D convolutional networks on coarse voxel grids~\cite{dai2017shape,tatarchenko2017octree,han2017high,varley2017shape,yang2018dense,shin2018pixels,xie2019pix2vox} and later expanded to point clouds~\cite{yang2018foldingnet,yuan2018pcn,tchapmi2019topnet} and triangle meshes~\cite{groueix2018papier,wang2018pixel2mesh}. More recently, implicit function representations using signed-distance~\cite{park2019deepsdf,xu2019disn} or binary occupancy~\cite{mescheder2019occupancy,peng2020convolutional,chibane2020implicit,humt2023shape,humt2023combining} fields have gained traction due to their simple training objective and strong representation power.

\noindent\textbf{Generative shape modeling.} Learning to fit distributions to shapes has followed a similar trajectory, from voxel~\cite{wu20153d,choy20163d,stutz2018learning}, point~\cite{achlioptas2018learning,yang2019pointflow,yu2021pointr} and mesh~\cite{liu2023meshdiffusion} to implicit~\cite{chen2018learning,chen2019unpaired,wu2020multimodal,zhang2021unsupervised,luo2021diffusion,zheng2022sdf,gao2022get3d,zheng2023locally,yan2022shapeformer,mittal2022autosdf,zhang20223dilg,zhang20233dshape2vecset,shue20233d,cheng2023sdfusion,chou2023diffusion,cui2024neusdfusion} representations.
These methods can be further demarcated along data~\cite{wu20153d,choy20163d,achlioptas2018learning,stutz2018learning,chen2018learning,yang2019pointflow,chen2019unpaired,wu2020multimodal,zhang2021unsupervised,luo2021diffusion,liu2023meshdiffusion,zheng2022sdf,gao2022get3d,zheng2023locally} or latent~\cite{yu2021pointr,yan2022shapeformer,mittal2022autosdf,zhang20223dilg,zhang20233dshape2vecset,shue20233d,cheng2023sdfusion,chou2023diffusion,cui2024neusdfusion} space generative modeling and into diffusion~\cite{luo2021diffusion,vahdat2022lion,zheng2023locally,zhang20233dshape2vecset,shue20233d,chou2023diffusion,cui2024neusdfusion} or autoregressive~\cite{yan2022shapeformer,mittal2022autosdf,zhang20223dilg} training paradigms.

\noindent\textbf{Single-view 3D reconstruction.} While closely related to shape completion, 3D reconstruction involves the additional challenge of transferring information from 2D to 3D. Due to its relevance and despite its complexity, it has attracted great attention among both discriminative~\cite{tatarchenko2017octree,kato2018neural,wang2018pixel2mesh,chen2018learning,shin2018pixels,xu2019disn,xie2019pix2vox} and generative~\cite{choy20163d,liu2023meshdiffusion,gao2022get3d,mittal2022autosdf,cui2024neusdfusion} methods.

\noindent\textbf{Shape Completion.} Obtaining the full 3D geometry from a partial, potentially degraded observation remains a significant challenge but has advanced significantly through both discriminative~\cite{dai2017shape,han2017high,varley2017shape,yang2018dense,yuan2018pcn,tchapmi2019topnet,humt2023shape,humt2023combining} and generative~\cite{wu20153d,stutz2018learning,chen2019unpaired,wu2020multimodal,yu2021pointr,zhang2021unsupervised,mittal2022autosdf,cheng2023sdfusion,chou2023diffusion,cui2024neusdfusion} modeling paradigms. Some additional works mention but do not focus on shape completion~\cite{achlioptas2018learning,park2019deepsdf,peng2020convolutional,yan2022shapeformer,zhang20233dshape2vecset}. Most works that focus on shape completion provide some quantitative evaluation~\cite{dai2017shape,han2017high,varley2017shape,stutz2018learning,yang2018dense,yuan2018pcn,tchapmi2019topnet,yu2021pointr,humt2023shape,humt2023combining,cheng2023sdfusion,chou2023diffusion,cui2024neusdfusion}, but rely either exclusively on global dataset statistics~\cite{wu2020multimodal,mittal2022autosdf,cheng2023sdfusion,chou2023diffusion,cui2024neusdfusion} or instance-level reconstruction quality~\cite{dai2017shape,han2017high,varley2017shape,stutz2018learning,yang2018dense,yuan2018pcn,tchapmi2019topnet,yu2021pointr,humt2023shape,humt2023combining}. A direct comparison between generative and discriminative models for shape completion is still missing.

The shape completion task is not clearly defined and is therefore used to mean different things in different works. Most works simply remove parts of the input using a cutting plane or volume. Some works render depth images~\cite{dai2017shape,han2017high,varley2017shape,stutz2018learning,yang2018dense,chen2019unpaired,zhang2021unsupervised,humt2023shape,humt2023combining} few of which additionally add some noise to the projected point cloud~\cite{humt2023shape,humt2023combining}. Except for~\cite{shin2018pixels,tatarchenko2019single,humt2023shape,humt2023combining}, the vast majority of works train in an object-centered coordinate system~\cite{tatarchenko2019single} instead of in camera--i.e. \emph{view-centered}--coordinates which significantly simplifies the task.

\section{Method}%
\label{sec:method}

\begin{figure*}[htbp]
  \centering
  \includegraphics[width=\textwidth]{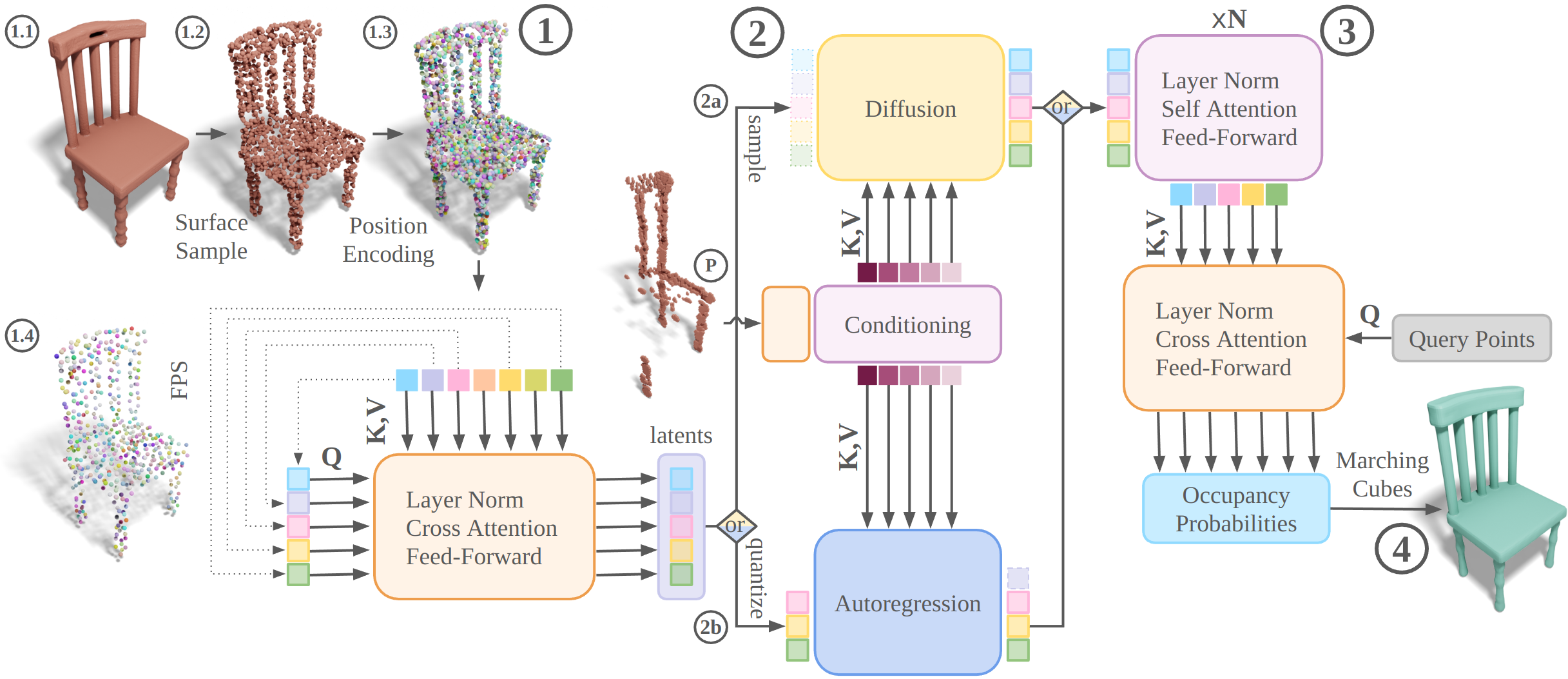}
  \caption{\textbf{Generative shape completion.} (1) Given an {\color{BrickRed}\textbf{input point cloud}} (1.2) sampled from the \textcolor{BrickRed}{\textbf{surface of an object}} (1.1), we apply a positional encoding (1.3) and aggregate the entire point cloud into a farthest-point-sampled (FPS) set (1.4) as in~\citet{zhang20233dshape2vecset}, which we additionally passed through a feed-forward network to form a latent code. (2) We then model these latents \emph{either} as a diagonal, multivariate Gaussian (2a) \emph{or} quantize them into a fixed-sized codebook (2b) forming our (VQ-)VAE encoder and train a diffusion \emph{or} autoregressive model on top, respectively. For shape completion, we condition the generative model on the encoding of a \textcolor{BrickRed}{\textbf{partial view}} (P) using a pre-trained feature extractor, which shares the overall architecture of the VAE. (3) We then predict occupancy probabilities through cross-attention between query points and latents sampled from the latent generative model, processed by $N$ Transformer encoder layers, forming the VAE decoder. (4) Optionally, a \textcolor{JungleGreen}{\textbf{mesh}} can be extracted using the \emph{Marching Cubes} algorithm. During inference, we discard the VAE encoder and sample latent codes \emph{either} autoregressively \emph{or} via denoising of samples drawn from a standard normal distribution.}
  \label{fig:method}
\end{figure*}

\textbf{Preliminaries.} Given a 3D object shape represented by a point cloud $x=\{x_i\in\mathbb{R}^3\}_{i=1}^N\in\mathcal{X}$ we train a VAE $f_\theta$ to predict the binary occupancy probability for any point $p\in\mathbb{R}^3$ as $f_\theta:\mathbb{R}^3\times\mathcal{X}\rightarrow[0,1]$ which is equivalent to the \emph{discriminative} training objective 
\begin{equation}
  \hat{y}=p_\theta(y=1\mid p,x)
\end{equation}
where $y\in\{0,1\}$ is the occupancy label for point $p$ and $\hat{y}\in[0,1]$ is the predicted occupancy probability. The VAE consists of an encoder $E$ that maps the input point cloud to a latent code $z=E(x)$ and a decoder $D$ that tries to map the latent code back to the input space, giving $\hat{x}=D(E(x))$.

Once the VAE is trained, we fit a generative prior $G$ on its latent distribution $p(z)$ to increase its expressiveness. We can further condition $G$ on signal $c$ during training to control the generation process. We train both a diffusion~\cite{ho2020denoising} model,
\begin{equation}
  p_\phi(z_{0:T}\mid c)=p(z_T)\prod_{t=1}^T p_\phi(z_{t-1},c)
\end{equation}
and an autoregressive~\cite{vaswani2017attention} model,
\begin{equation}
  p_\phi(z\mid c)=\prod_{i=1}^L p_\phi(z_{<i},c)
\end{equation}, where $T$ is the number of (de)noising steps and $L$ the number of autoregressive steps.

\noindent\textbf{Model architecture.} We build on~\citet{zhang20233dshape2vecset} for the VAE and diffusion model architectures. As shown in Fig.~\ref{fig:method}, the VAE encoder ingests positional encoded, sampled surface points and cross-attends~\cite{vaswani2017attention} (also sometimes referred to as \emph{encoder-decoder attention}) to farthest-point-sampled (FPS) \emph{queries} to encode the surface points into a fixed-length latent set. From this latent set, a diagonal Gaussian parameterization is predicted for the VAE while being quantized into fixed codebook entries in the VQ-VAE~\cite{van2017neural} case, as required for autoregressive training. The sampled (VAE) or quantized (VQ-VAE) latent code is then processed by multiple Transformer~\cite{vaswani2017attention} encoder layers with layer norm, self-attention, and feed-forward components. Finally, the occupancy probability for $p$ is predicted through cross-attention between positional encoded point coordinates and the latent code. We refer to~\citet{zhang20233dshape2vecset} for further details.

We make the following changes to the VAE architecture of~\citet{zhang20233dshape2vecset}: (1) We use the original NeRF~\cite{mildenhall2021nerf} positional encoding for both the surface and occupancy points. (2) We add a layer-normalization and feed-forward component to the input encoding stage. (3) We use multi-headed attention~\cite{vaswani2017attention} throughout the entire model. (4) We half the input dimension of all GeGLU~\cite{shazeer2020glu} activations. These changes allow us to train a VAE of one-third the size of the original while achieving the same performance (Tab.~\ref{tab:model_size}).

Despite a large codebook as suggested in~\citet{rombach2022high} and various improvements to VQ-VAE training from the literature like K-means initialization~\cite{zeghidour2021soundstream} and compression of the codebook dimension~\cite{yu2021vector} which indeed increase reconstruction quality, we are unable to match the performance of the continuous VAE (Tab.~\ref{tab:recon_quality}). We found codebook sampling~\cite{lee2022autoregressive} and regularization~\cite{yu2021vector}, expiring of stale codes~\cite{zeghidour2021soundstream} and Finite Scalar Quantization~\cite{mentzer2023finite} as well as Lookup Free Quantization~\cite{yu2023language} to be ineffective (ablations can be found in the supplementary material).

For unconditional generative training, both the diffusion and autoregressive models share the same Transformer encoder design. In their conditional configuration, all layers are replaced by Transformer decoder blocks, which add a cross-attention component. The autoregressive model uses \emph{causal} self-attention. As an alternative to conditioning via cross-attention, we can prepend the conditioning vector to the latent code (Tab.~\ref{tab:cond_type}). The diffusion model uses adaptive layer normalization~\cite{perez2018film} for time-step conditioning.

\noindent\textbf{Training.} We train all models in mixed precision using the Adam~\cite{kingma2014adam} optimizer with a linear warmup, cosine annealing learning rate schedule peaking at $0.0001$ and effective batch size of $256$ on $4$-$8$ NVIDIA A100 80GB GPUs for $800$-$2000$ epochs. We use weight decay of $0.005$, exponential moving average over weights and gradient clipping. We found the former to benefit diffusion model performance and the latter being crucial for stable (VQ-)VAE training. During the auto-encoding stage, we augment the inputs by adding Gaussian noise to the surface points and independently randomly scale all axes by up to $20\%$. Contrary to~\citet{zhang20233dshape2vecset}, we do not use this type of augmentation during training of the latent generative model to prevent the generation of distorted shapes during inference.

\section{Experiments}%
\label{sec:experiments}

This section comprises four parts: We begin by discussing evaluation metrics, then validate our models' reconstruction and generative modeling performance. Next, we assess shape completion capabilities under increasing complexity and realism. Finally, we conduct ablation studies examining how various design choices affect overall model performance.

All experiments utilize the ShapeNet (v1) dataset~\cite{chang2015shapenet}, unless otherwise specified, with training data generated following the approach of~\citet{humt2023combining}.

\subsection{Metrics}%
\label{sub:metrics}

As alluded to in the introduction, many evaluation metrics for reconstruction and generative modeling have been proposed, and no consensus on their relative importance has been reached. We, therefore, evaluate our models across a wide range of metrics to provide a comprehensive view of their performance.

\noindent\textbf{Instance-level.} To evaluate the reconstruction quality, we rely on (volumetric) \emph{Intersection-over-Union} (IoU) (if applicable) and bidirectional L1 \emph{Chamfer Distance} (CD), scaled following~\citet{mescheder2019occupancy}. We further make use of \emph{F1-score} as well as \emph{Precision} and \emph{Recall}, also referred to as \emph{accuracy} and \emph{completeness} in~\citet{tatarchenko2019single}. IoU can only be evaluated for watertight meshes, but we opt to evaluate against the original meshes from ShapeNet to facilitate reproduction and comparison and for consistency with the generative modeling evaluation.

\noindent\textbf{Set-level.} All of these metrics measure \emph{instance-level} performance, as opposed to the following metrics most commonly used for evaluating the generative quality, which measure \emph{set-level} or \emph{global} performance.

The earliest metrics for evaluating generative models are \emph{Minimum-Matching-Distance} (MMD) and \emph{Coverage} (COV)~\cite{achlioptas2018learning} which we retire in favor of \emph{Leave-One-Out 1-Nearest-Neighbor Accuracy} (1-NNA), proposed to alleviate the shortcomings of MMD and COV~\cite{yang2019pointflow}. Some works also use \emph{Edge Count Difference} (ECD)~\cite{ibing20213d} as well as \emph{Total Mutual Distance} (TMD) and \emph{Unidirectional Hausdorff Distance} (UHD)~\cite{wu2020multimodal} which we found to be less informative and refer to the appendix.

More recent additions are Fréchet~\cite{shu20193d} and Kernel~\cite{zhang20233dshape2vecset} Pointcloud Distance (KPD, FPD), which compute the Fréchet and Kernel distance between point features extracted from the generated and ground truth surface points. These are highly informative but rely on a pre-trained feature extractor, which each work redefines and retrains, making comparison impossible. We reuse the features of our VAE trained on the reconstruction task, which we find to be more informative than the commonly used features from models trained on point cloud classification. Furthermore, this not only frees us from training yet another model but also provides a close to normally distributed feature space, an implicit assumption of the Fréchet distance. Following prior work~\cite{achlioptas2018learning,yang2019pointflow,zheng2022sdf,zhang20233dshape2vecset}, these set-level metrics are computed on the test split.

Finally, we also evaluate the perceptual quality of the generated shapes using the (shading-image-based) \emph{Fréchet Inception Distance} (FID)~\cite{heusel2017gans} and \emph{Kernel Inception Distance} (KID)~\cite{binkowski2018demystifying} which measure the distance between the feature distributions of images of generated and real objects, rendered from uniformly sampled viewpoints. Here, we additionally employ CLIP~\cite{radford2021learning} features from a Vision Transformer~\cite{dosovitskiy2020image} as proposed in~\citet{kynkaanniemi2022role} and shown to align better with human perception than Inception features (results for this metric can be found in the appendix).

\citet{sajjadi2018assessing} show how FID can be decomposed into \emph{Precision} and \emph{Recall} of which we use the improved version by \citet{kynknniemi2019improved} relying on k-NN instead of k-means clustering. We propose to employ the same decomposition for KPD. \citet{ferjad2020reliable} claim an even better decomposition into \emph{Density} and \emph{Coverage} exists, but due to lack of adoption, these are provided in the appendix.

Again, following prior convention, we evaluate FID and its decompositions as well as KID on the train split.

\subsection{Results}%
\label{sub:results}

\label{sub:reconstruction}
\textbf{Reconstruction.} The reconstruction quality of the VAE determines the upper bound for the latent generative model performance. As shown in Tab.~\ref{tab:recon_quality}, first row, our VAE achieves comparable performance to the current SOTA~\cite{zhang20233dshape2vecset} on watertight meshes, but due to differences in training data, no direct comparison can be made. The VQ-VAE (second row) falls short of the VAE but performs reasonably well for a large and diverse dataset such as ShapeNet.
We also include the performance on the original ShapeNet meshes in the table's lower half (3rd and 4th row) for reference and future comparison. Interestingly, while the models are well calibrated on the watertight meshes, achieving similar precision and recall, on the original meshes, recall is lacking behind significantly, which we attribute to loss of (interior) detail during the watertightening process.

\begin{table}[htbp]
\centering
\setlength{\tabcolsep}{4pt}
\begin{tabular}{lS[round-precision=3,table-format=1.3]S[table-format=2.2]S[table-format=2.2]S[table-format=2.2]}
  \toprule
  & {Chamfer $\downarrow$} & {F1 $\uparrow$} & {Precision $\uparrow$} & {Recall $\uparrow$} \\
  \midrule
  VAE & \bfseries 0.0317805 & \bfseries 98.3313 & \bfseries 98.6234 & \bfseries 98.1279 \\
  VQ-VAE & 0.06865 & 89.3320 & 89.3400 & 89.8313 \\
  \midrule
  VAE & \bfseries 0.0911306 & \bfseries 77.187 & \bfseries 82.6042 & \bfseries 74.528 \\
  VQ-VAE & 0.11609 & 70.5324 & 74.4680 & 69.0812 \\
  \bottomrule
\end{tabular}
\caption{Reconstruction quality; class average. Upper half shows performance on watertight meshes, lower half on original meshes.}
\label{tab:recon_quality}
\end{table}

\label{sub:generative_modeling}
\noindent\textbf{Generative Modeling.} To validate the performance of the latent generative training, we compare our class-conditional LDM against two recent SOTA baselines, LAS-Diffusion~\cite{zheng2023locally} (LAS-Dif.) and 3DShape2VecSet~\cite{zhang20233dshape2vecset} (3DS2VS) on the same subset of classes. We use the provided model checkpoints, as retraining these models incurs a significant computational overhead. The results are shown in Tab.~\ref{tab:class_cond_test}.

Due to differences in training data and procedures, we are able to outperform the superior
3DShape2VecSet baseline across all metrics while sharing the overall model architecture. All models show much higher precision than recall, indicating paths toward future improvement.

\begin{table}[htbp]
\centering
\setlength{\tabcolsep}{2.5pt}
\sisetup{
  round-precision=2
}
\begin{tabular}{llS[table-format=2.2]HS[table-format=3.2]S[table-format=3.2]S[table-format=2.2]S[table-format=2.2]HH}
  \toprule
  & {} & {1-NNA$\downarrow$} & {ECD$\downarrow$} & {FPD$\downarrow$} & {KPD$\downarrow$} & {Prec.$\uparrow$} & {Rec.$\uparrow$} & {Dens.$\uparrow$} & {Cov.$\uparrow$} \\
  \midrule
  \multirow{3}{*}{\rotatebox{90}{\textbf{Chair}}}
    & LAS-Diff. & 59.08419497784343 & 80.49428184979755 & 99.17017632099714 & 9.312835463831902 & \bfseries 96.89807976366323 & 63.36779911373708 & \bfseries 1.3031019202363368 & 0.8892171344165436 \\
    & 3DS2VS & 58.93648449039882 & 25.97728184206988 & 94.01398448338341 & 7.1597585662331475 & 85.67208271787297 & \bfseries 77.10487444608567 & 0.7878877400295421 & 0.8286558345642541 \\
    & Ours & \bfseries 58.49335302806499 & \bfseries 7.057903294490758 & \bfseries 89.58658797523253 & \bfseries 6.97410854214546 & 95.56868537666174 & 60.70901033973413 & 1.3028064992614476 & \bfseries 0.9069423929098966 \\
  \midrule
  \multirow{3}{*}{\rotatebox{90}{\textbf{Plane}}}
    & LAS-Diff. & 82.67326732673267 & 164.4906044974176 & 257.6618045508026 & 34.79479911264624 & 75.0 & 11.386138613861387 & 0.43514851485148515 & 0.3217821782178218 \\
    & 3DS2VS & 69.67821782178217 & 37.3497194645217 & 165.00836229770084 & 22.3454074706489 & 68.31683168316832 & \bfseries 34.15841584158416 & 0.4618811881188119 & 0.4628712871287129 \\
    & Ours & \bfseries 69.05940594059405 & \bfseries 10.22629097498548 & \bfseries 139.68486469301092 & \bfseries 17.054163521795438 & \bfseries 84.15841584158416 & 30.94059405940594 & \bfseries 0.7217821782178219 & \bfseries 0.556930693069307 \\
  \midrule
  \multirow{3}{*}{\rotatebox{90}{\textbf{Car}}}
    & LAS-Diff. & 86.3150867823765 & \bfseries 482.99944047975106 & 99.01572031076898 & \bfseries 16.26722705225314 & \bfseries 62.61682242990654 & \bfseries 55.67423230974633 & 0.27476635514018694 & \bfseries 0.3591455273698264\\
    & 3DS2VS & 91.0547396528705 & 2035.9806491919655 & 170.99114959950953 & 27.70528120088605 & 60.88117489986649 & 39.65287049399199 & 0.22429906542056074 & 0.17623497997329773 \\
    & Ours & \bfseries 82.1762349799733 & 538.104533054052 & \bfseries 84.73930947206367 & 16.61974036928133 & 59.41255006675568 & 48.19759679572764 & \bfseries 0.27716955941255006 & 0.3030707610146863 \\
  \midrule
  \multirow{3}{*}{\rotatebox{90}{\textbf{Table}}}
    & LAS-Diff. & 55.35294117647059 & 135.57997966311905 & 158.87359836247379 & 19.95190206950745 & 94.58823529411765 & \bfseries 72.11764705882353 & 0.9985882352941178 & 0.8635294117647059 \\
    & 3DS2VS & 56.76470588235294 & 19.636860658267743 &  148.09748139253043& 15.082496804238199 & 92.47058823529412 & 71.52941176470589 & 0.9272941176470589 & 0.8294117647058824 \\
    & Ours & \bfseries 53.70588235294118 & \bfseries 15.702286851771053 & \bfseries 128.35344985256415 & \bfseries 9.533235844308027 & \bfseries 96.70588235294117 & 67.64705882352942 & \bfseries 1.203764705882353 & \bfseries 0.8717647058823529 \\
  \midrule
  \multirow{3}{*}{\rotatebox{90}{\textbf{Rifle}}}
    & LAS-Diff. & 77.42616033755274 & 180.3133876986132 & 693.8384515518387 & 115.62660763954709 & \bfseries 96.62447257383966 & 34.59915611814346 & 0.5341772151898735 & 0.21518987341772153 \\
    & 3DS2VS & \bfseries 66.03375527426161 & 38.638367919480075 & 418.0064691086095 & 57.748641784765674 & 91.9831223628692 & \bfseries 50.63291139240507 & 0.7147679324894516 & 0.4219409282700422 \\
    & Ours & 70.46413502109705 & \bfseries 11.755426656071002 & \bfseries 347.778028072869 & \bfseries 52.889068495706994 & 95.78059071729957 & 30.80168776371308 & \bfseries 0.8067510548523207 & \bfseries 0.4472573839662447 \\
  \toprule
  \multirow{3}{*}{\rotatebox{90}{\textbf{Mean}}}
    & LAS-Diff. & 72.17033012019519 & 208.7755388377397 & 261.71195021937626 & 39.190674267557156 & 85.14552201230542 & 47.428994642862354 &  0.7091564481424 & 0.5297728250373238 \\
    & 3DS2VS & 68.49358062433322 & 431.516575815261 & 199.22348937634675 & 26.00831716535439 & 79.86475997981421 & \bfseries 54.61569678775455 & 0.623226008741085 & 0.5438229589284378 \\
    & Ours & \bfseries 66.77980226453412 & \bfseries 116.56928816627405 & \bfseries 158.02844801314805 & \bfseries 20.61406335464745 & \bfseries 86.32522487104847 & 47.659189556422044 & \bfseries 0.8624547995252986 & \bfseries 0.6171931873684975 \\
  \bottomrule
\end{tabular}%
\caption{Comparison of \emph{class-conditional} generative models.}
\label{tab:class_cond_test}
\end{table}

We then proceed to compare our unconditional LDM and AR models. According to Tab.~\ref{tab:uncond_gen}, the AR model is outperformed by the LDM, which we attribute to the superior reconstruction quality of the VAE, as established in the previous section.

\begin{table}[htbp]
\centering
\begingroup
\sisetup{table-format=2.2}
\begin{tabular}{lBS}
  \toprule
  & {\normalfont Diffusion (VAE)} & {AR (VQ-VAE)} \\
  \midrule
  FID $\downarrow$ & 32.61765585013054 & 35.75992129187906 \\
  KID $\times10^3\downarrow$ & 12.998296084959962 & 13.165328860610611 \\
  Precision $\uparrow$ & 50.27036082474227 & 50.25940721649484 \\
  Recall $\uparrow$ & 48.079123711340205 & 42.07757731958763 \\
  \bottomrule
\end{tabular}
\endgroup
\caption{Comparison of diffusion and autoregressive \emph{unconditional} generative shape modeling on continuous (VAE) and discrete (VQ-VAE) latents.}
\label{tab:uncond_gen}
\end{table}

To test this hypothesis, we train both a class-conditional LDM and AR model on the \emph{same} discrete VQ-VAE latent space. This setup uses an embedding of the class labels as conditioning information $c$. As evident from Tab.~\ref{tab:class_cond_gen}, the AR model is able to outperform the LDM in this setting. For reference, we also include the results of the class-conditional LDM trained on the continuous VAE latents.

\begin{table}[htbp]
\centering
\begingroup
\sisetup{table-format=2.2}
\begin{tabular}{lSS|S}
  \toprule
  & \multicolumn{2}{c}{{VQ-VAE}} & {VAE} \\
  \midrule
  & {Diffusion} & {Autoregressive} & {Diffusion} \\
  \midrule
  FID $\downarrow$ & 42.98419957592065 & \bfseries 33.58201232088938 & 30.01671346063884 \\
  KID$\times10^3\downarrow$ & 18.032472851601606 & \bfseries 12.049536292917926 & 11.158433145895895 \\
  Precision $\uparrow$ & 38.585567010309274 & \bfseries 51.60605670103091 & 53.93595360824742 \\
  Recall $\uparrow$ & 37.975128865979385 & \bfseries 43.507474226804116 & 46.881572164948454 \\
  \midrule
  1-NNA $\downarrow$ & 67.93323863636364 & \bfseries 66.53941761363636 & 65.01242897727273 \\
  FPD $\downarrow$ & 80.38031029393801 & \bfseries 77.50991482330028 & 73.03440617352317 \\
  KPD $\downarrow$ & \bfseries 4.302245640991602 & 5.140275468205307 & 4.425934648945596 \\
  Precision $\uparrow$ & \bfseries 92.16974431818182 & 91.61931818181818 & 91.51278409090909 \\
  Recall $\uparrow$ & 54.19034090909091 & \bfseries 60.86647727272727 & 59.99644886363636 \\
  \bottomrule
\end{tabular}
\endgroup
\caption{Comparison of diffusion and autoregressive \emph{class-conditional} generative shape modeling on the same latent space.}
\label{tab:class_cond_gen}
\end{table}

\label{sub:shape_completion}
\noindent\textbf{Shape Completion.} We now come to the main results of this work, comparing the discriminative and generative approach on the shape completion task. Our discriminative model architecture is identical to the VAE used for shape auto-encoding, except for the variational part and the fact that the input is now a partial view of the object. The encoder of the trained discriminative model is repurposed as feature extractor to the latent generative model to provide highly informative conditioning information. We tried training a dedicated feature extractor on the classification task but found this to result in worse performance (Tab.~\ref{tab:cond_type}).

The simplest task we consider is shape completion from a rendered depth image in object-centric coordinates. Due to self-occlusions, this is still significantly more challenging than random removal of parts of the object, as is common practice. Contrary to unconditional and class-conditional generation, shape completion is only evaluated on the test split, as we are interested in generalization to novel instances instead of faithfully capturing the underlying data distribution. In this simplified setup, the discriminative model is able to slightly outperform the generative model on both set-level (upper part) and instance-level (lower part) metrics (Tab.~\ref{tab:depth_cond_test}). Following~\citet{tatarchenko2019single}, this is to be expected, as the discriminative model can bypass the complex shape completion task and learn the more straightforward retrieval task instead.

\begin{table}[htbp]
\sisetup{
  round-precision=3,
  table-format=2.3
}
\centering
\begin{tabular}{lSS}
  \toprule
  & {\color{darkgray}\textbf{Discriminative}} & {\color{JungleGreen}\textbf{Generative}} \\
  \midrule
  1-NNA $\downarrow$ & \bfseries 30.45099431818182 & 30.806107954545453 \\
  FPD $\downarrow$ & \bfseries 71.12558537426435 & 71.7820624472024 \\
  KPD $\downarrow$ & \bfseries 5.622214782548493 & 6.1149716704223955 \\
  Precision $\uparrow$ & 93.92755681818182 & \bfseries 94.85085227272727 \\
  Recall $\uparrow$ & \bfseries 77.18394886363636 & 76.38494318181818 \\
  \midrule
  Chamfer $\downarrow$ & \bfseries 0.117832 & 0.121743 \\
  F1 $\uparrow$ & \bfseries 70.6812 & 69.0981 \\
  Precision $\uparrow$ & \bfseries 74.7947 & 72.4853 \\
  Recall $\uparrow$ & \bfseries 69.2264 & 68.1494 \\
  \bottomrule
\end{tabular}
\caption{{\color{JungleGreen}\textbf{Generative}} vs. {\color{darkgray}\textbf{discriminative}} shape completion from a single depth image in object-centric coordinates.}
\label{tab:depth_cond_test}
\end{table}

Moving on to shape completion in camera coordinates (Tab.~\ref{tab:depth_cond_cam_test}), the results from the previous tasks are reversed for the set-level metrics. Now, the generative model appears slightly better than the discriminative model, which can no longer entirely rely on the retrieval shortcut. Still, the discriminative model has a slight edge in instance-level performance.

To understand why, recall that discriminative models are forced to predict the best \emph{average} result when faced with ambiguous inputs, whereas generative models when only queried once, can and will predict a single, plausible result, which is not necessarily as close to the ground truth as the average. To test this hypothesis, we move on to the final, most complex shape completion task: the completion of noisy depth images (in camera coordinates) as captured by widely available RGB-D sensors like the \emph{Microsoft Kinect}.

\begin{table}[htbp]
\sisetup{
  round-precision=3,
  table-format=2.3
}
\centering
\begin{tabular}{lSS}
  \toprule
  & {\color{darkgray}\textbf{Discriminative}} & {\color{JungleGreen}\textbf{Generative}} \\
  \midrule
  1-NNA $\downarrow$ & 31.480823863636365 & \bfseries 31.099076704545453 \\
  FPD $\downarrow$ & 74.81695053823796 & \bfseries 70.26922976942728 \\
  KPD $\downarrow$ & 6.5404486539061155 & \bfseries 5.892627758115871 \\
  Precision $\uparrow$ & 92.13423295454546 & \bfseries 92.27627840909091 \\
  Recall $\uparrow$ & 77.55681818181818 & \bfseries 77.61008522727273 \\
  Density $\uparrow$ & 0.8453125 & \bfseries 0.9204190340909092 \\
  Coverage $\uparrow$ & 0.7823153409090909 & \bfseries 0.7950994318181818 \\
  \midrule
  Chamfer $\downarrow$ & \bfseries 0.125052 & 0.128004 \\
  F1 $\uparrow$ & \bfseries 69.0699 & 68.0809 \\
  Precision $\uparrow$ & \bfseries 74.1737 & 72.6689 \\
  Recall $\uparrow$ & \bfseries 66.4754 & 65.9115 \\
  \bottomrule
\end{tabular}
\caption{{\color{JungleGreen}\textbf{Generative}} vs. {\color{darkgray}\textbf{discriminative}} shape completion from a single depth image in camera coordinates}
\label{tab:depth_cond_cam_test}
\end{table}

\begin{figure}[htbp]
    \centering
    \begin{subfigure}{0.25\linewidth}
        \includegraphics[width=\textwidth]{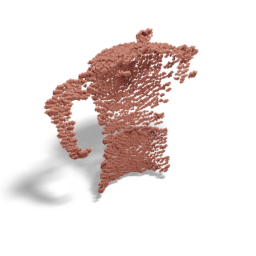}%
    \end{subfigure}%
    \begin{subfigure}{0.25\linewidth}
        \includegraphics[width=\textwidth]{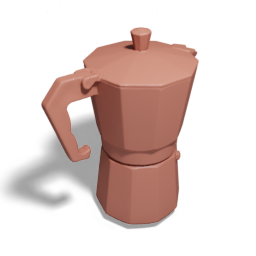}%
    \end{subfigure}%
    \begin{subfigure}{0.25\linewidth}
        \includegraphics[width=\textwidth]{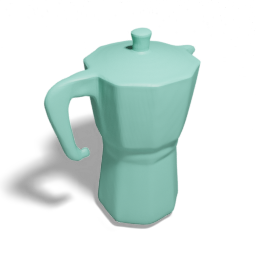}%
    \end{subfigure}%
    \begin{subfigure}{0.25\linewidth}
        \includegraphics[width=\textwidth]{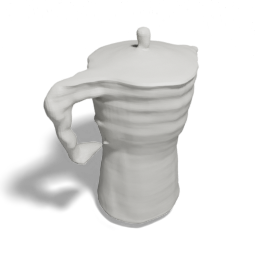}%
    \end{subfigure}%

    \begin{subfigure}{0.25\linewidth}
        \includegraphics[width=\textwidth]{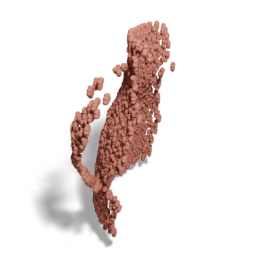}%
    \end{subfigure}%
    \begin{subfigure}{0.25\linewidth}
        \includegraphics[width=\textwidth]{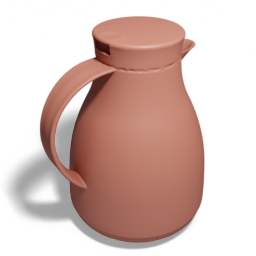}%
    \end{subfigure}%
    \begin{subfigure}{0.25\linewidth}
        \includegraphics[width=\textwidth]{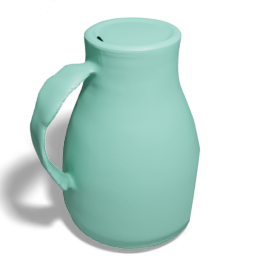}%
    \end{subfigure}%
    \begin{subfigure}{0.25\linewidth}
        \includegraphics[width=\textwidth]{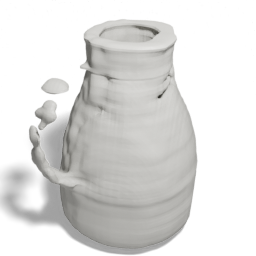}%
    \end{subfigure}%

    \begin{subfigure}{0.25\linewidth}
        \includegraphics[width=\textwidth]{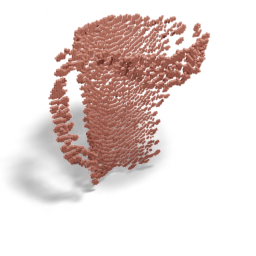}%
    \end{subfigure}%
    \begin{subfigure}{0.25\linewidth}
        \includegraphics[width=\textwidth]{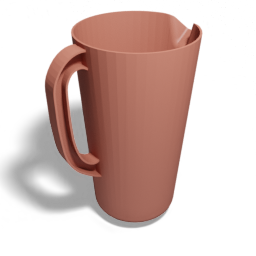}%
    \end{subfigure}%
    \begin{subfigure}{0.25\linewidth}
        \includegraphics[width=\textwidth]{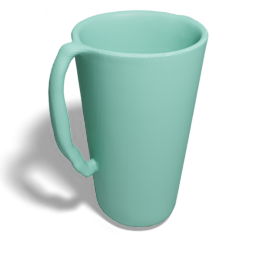}%
    \end{subfigure}%
    \begin{subfigure}{0.25\linewidth}
        \includegraphics[width=\textwidth]{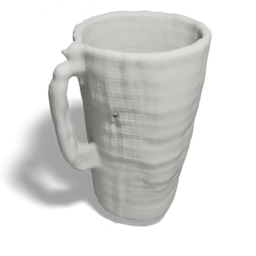}%
    \end{subfigure}%

    \begin{subfigure}{0.24\linewidth}
        \includegraphics[width=\textwidth]{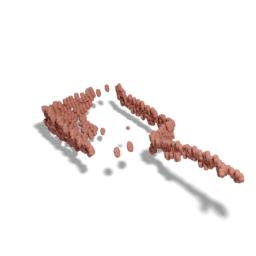}%
    \end{subfigure}%
    \begin{subfigure}{0.24\linewidth}
        \includegraphics[width=\textwidth]{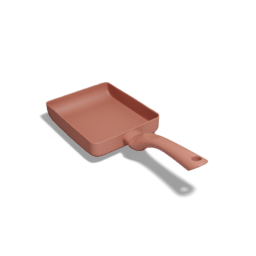}%
    \end{subfigure}%
    \begin{subfigure}{0.24\linewidth}
        \includegraphics[width=\textwidth]{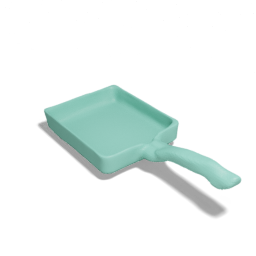}%
    \end{subfigure}%
    \begin{subfigure}{0.24\linewidth}
        \includegraphics[width=\textwidth]{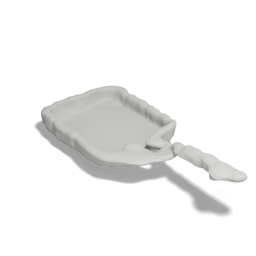}%
    \end{subfigure}%

    \begin{subfigure}{0.24\linewidth}
        \includegraphics[width=\textwidth]{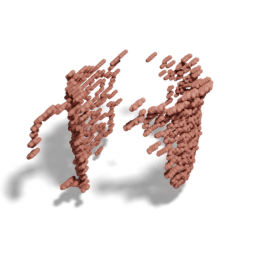}%
    \end{subfigure}%
    \begin{subfigure}{0.24\linewidth}
        \includegraphics[width=\textwidth]{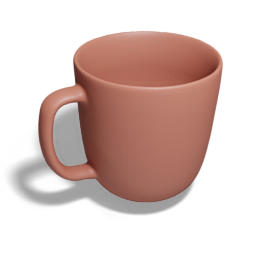}%
    \end{subfigure}%
    \begin{subfigure}{0.24\linewidth}
        \includegraphics[width=\textwidth]{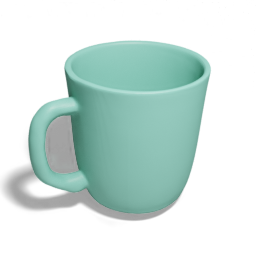}%
    \end{subfigure}%
    \begin{subfigure}{0.24\linewidth}
        \includegraphics[width=\textwidth]{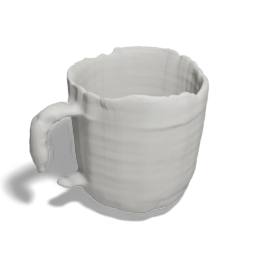}%
    \end{subfigure}%
  \caption{Examples from the \emph{Automatica/YCB} dataset. Left to right: {\color{BrickRed}\textbf{input}}, {\color{BrickRed}\textbf{ground truth}}, {\color{JungleGreen}\textbf{generative}} (best), {\color{darkgray}\textbf{discriminative}}.}
    \label{fig:ycb_cond_test}
\end{figure}

Instead of generating a single completion, which goes against a generative model's actual benefit and strength, we now instead produce $10$ completions per input and pick the one with the highest F1-score. We argue that this is the correct way to assess the generative model's upper-bound performance, as we are interested in its ability to produce not just plausible but also more accurate results than a discriminative model when faced with ambiguous inputs. Tab.~\ref{tab:kinect_cond_test} confirms our hypothesis in which the generative model (G) now consistently outperforms the discriminative model (D) by a large margin across all metrics. This effect can also be observed in Fig.~\ref{fig:hero} and~\ref{fig:ycb_cond_test} where the generative model produces always plausible and, in the best case, also more accurate completions than the discriminative model.

\begin{table*}[htbp]
  \centering
  \setlength{\tabcolsep}{4.5pt}
  \sisetup{
    round-precision=2,
    table-format=2.2
  }
  \begin{tabular}{lSSHHS[round-precision=0,table-format=3.0]S[round-precision=0,table-format=3.0]SSHHHH|S[round-precision=3,table-format=1.3]S[round-precision=3,table-format=1.3]SSSSSS}
    \toprule
    & \multicolumn{2}{c}{{1-NNA$\downarrow$}} & \multicolumn{2}{H}{{ECD$\downarrow$}} & \multicolumn{2}{c}{{FPD$\downarrow$}} & \multicolumn{2}{c}{{KPD$\downarrow$}} & \multicolumn{2}{H}{{Prec.$\uparrow$}} & \multicolumn{2}{H}{{Rec.$\uparrow$}} & \multicolumn{2}{|c}{{CD$\downarrow$}} & \multicolumn{2}{c}{{F1$\uparrow$}} & \multicolumn{2}{c}{{Prec.$\uparrow$}} & \multicolumn{2}{c}{{Rec.$\uparrow$}} \\
    \midrule
    & {\color{darkgray}\textbf{D}} & {\color{JungleGreen}\textbf{G}} & {\color{darkgray}\textbf{D}} & {\color{JungleGreen}\textbf{G}} & {\color{darkgray}\textbf{D}} & {\color{JungleGreen}\textbf{G}} & {\color{darkgray}\textbf{D}} & {\color{JungleGreen}\textbf{G}} & {\color{darkgray}\textbf{D}} & {\color{JungleGreen}\textbf{G}} & {\color{darkgray}\textbf{D}} & {\color{JungleGreen}\textbf{G}} & {\color{darkgray}\textbf{D}} & {\color{JungleGreen}\textbf{G}} & {\color{darkgray}\textbf{D}} & {\color{JungleGreen}\textbf{G}} & {\color{darkgray}\textbf{D}} & {\color{JungleGreen}\textbf{G}} & {\color{darkgray}\textbf{D}} & {\color{JungleGreen}\textbf{G}} \\
    Chair & 38.25701624815362 & \bfseries 33.677991137370755 & 118.40910474143494 & \bfseries 95.36589650748068 & 299.77586420908483 & \bfseries 128.3521815246902 & 48.546258758311104 & \bfseries 16.26660255961808 & 80.35450516986706 & \bfseries 87.74002954209749 & 48.892171344165436 & \bfseries 85.81979320531757 & 0.395799 & \bfseries 0.32730 & 43.2726 & \bfseries 52.4714 & 47.1179 & \bfseries 56.6614 & 41.0942 & \bfseries 50.4970 \\
    Plane & 60.27227722772277 & \bfseries 50.99009900990099 & 119.06117707569913 & \bfseries 72.73906505499616 & 425.6293969581641 & \bfseries 213.86719934466237 & 62.18351376616974 & \bfseries 28.971572504498745 & 75.74257425742574 & \bfseries 85.39603960396039 & 14.85148514851485 & \bfseries 48.267326732673266 & 0.311524 & \bfseries 0.29014 & 47.0918 & \bfseries 55.6980 & 50.6998 & \bfseries 62.4971 & 45.3234 & \bfseries 51.8630 \\
    Car & 88.18424566088118 & \bfseries 74.6995994659546 & 1445.4602492654508 & \bfseries 394.88978063394995 & 172.11391390453423 & \bfseries 114.32870706500916 & 33.884359429709235 & \bfseries 17.852929726043726 & 27.369826435246997 & \bfseries 49.265687583444595 & 12.016021361815754 & \bfseries 54.47263017356475 & 0.282849 & \bfseries 0.25958 & 38.118 & \bfseries 47.6417 & 50.4374 & \bfseries 58.9671 & 31.1896 & \bfseries 40.9205 \\
    Table & 38.941176470588235 & \bfseries 35.94117647058824 & 35.82915101804676 & \bfseries 16.37833482094103 & 274.44469570192086 & \bfseries 134.06604916980314 & 31.130998642888883 & \bfseries 13.570355501565526 & 84.82352941176471 & \bfseries 95.76470588235294 & 64.11764705882353 & \bfseries 79.17647058823529 & 0.447081 & \bfseries 0.26435 & 48.0384 & \bfseries 57.1465 & 49.9848 & \bfseries 60.3705 & 48.2835 & \bfseries 56.1067 \\
    Rifle & 57.38396624472574 & \bfseries 54.21940928270043 & 82.89059439114324 & \bfseries 73.63435002288752 & 572.242164205503 & \bfseries 479.7763144782542 & 85.24054806676735 & \bfseries 69.67679097630563 & 75.10548523206751 & \bfseries 91.13924050632911 & \bfseries 40.92827004219409 & 33.755274261603374 & 0.550823 & \bfseries 0.54209 & 37.3875 & \bfseries 45.9956 & 41.2176 & \bfseries 55.5294 & 35.8422 & \bfseries 41.5188 \\
    \midrule
    Mean & 56.607736 & \bfseries 49.905655073303 & 360.330055 & \bfseries 130.60148540805108 & 348.841207 & \bfseries 214.07809031648384 & 52.197136 & \bfseries 29.267650253606337 & 68.679184 & \bfseries 81.8611406236369 & 36.161119 & \bfseries 60.298298992278866 & 0.397615 & \bfseries 0.336692 & 42.78166 & \bfseries 51.79064 & 47.8915 & \bfseries 58.8051 & 40.34658 & \bfseries 48.18120 \\
    \midrule
    All & 53.59552556818182 & \bfseries 48.52627840909091 & \bfseries 189.27667301350087 & 238.492222159629 & 203.88417518397864 & \bfseries 103.42393011717127 & 24.347399180570868 & \bfseries 9.257547429556073 & 77.16619318181818 & \bfseries 88.4765625 & 41.08664772727273 & \bfseries 70.73863636363636 & / & / & / & / & / & / & / & / \\
    \bottomrule
  \end{tabular}
  \caption{{\color{JungleGreen}\textbf{Generative}} ({\color{JungleGreen}\textbf G}) vs. {\color{darkgray}\textbf{discriminative}} ({\color{darkgray}\textbf D}) shape completion from a single Kinect depth image. Instance-level metrics (CD, F1, Prec., Rec.) are 'best-of-10' for the generative model, which is already competitive with the discriminative model at N=1 (See Tab.~\ref{tab:num_completions}).}
  \label{tab:kinect_cond_test}
\end{table*}

In a final experiment, we investigate the model performance under domain shift and evaluate both the discriminative and generative model on the \emph{Automatica/YCB} dataset by~\citet{humt2023combining}. The results in Tab.~\ref{tab:ycb_cond_test} show both the superior performance of our discriminative model over the equivalent model of~\cite{humt2023combining} which are still further improved upon by the generative model, as illustrated in Fig.~\ref{fig:ycb_cond_test}. While Chamfer distance remains unchanged for the discriminative model and slightly increases for the generative model, this metric is strongly effected by outliers and poor at distinguishing visual quality~\cite{achlioptas2018learning,liu2020morphing,wu2021density}. Qualitative results on real Kinect depth data can be found in the appendix.

\begin{table}[htbp]
\centering
\begin{tabular}{lS[round-precision=3,table-format=1.3]S[round-precision=2,table-format=2.2]S[round-precision=2,table-format=2.2]S[round-precision=2,table-format=2.2]}
  \toprule
  & {CD $\downarrow$} & {F1 $\uparrow$} & {Prec. $\uparrow$} & {Rec. $\uparrow$} \\
  \midrule
  \textbf{Kinect} \cite{humt2023combining} & 0.304548 & 43.3697 & 44.6928 & 42.851 \\
  \textcolor{darkgray}{\textbf{Discriminative}} & \bfseries 0.297257 & 45.9852 & 47.0179 & 45.8319 \\
  \textcolor{JungleGreen}{\textbf{Generative}} & 0.345843 & \bfseries 52.9221 & \bfseries 54.2291 & \bfseries 52.4319 \\
  \bottomrule
\end{tabular}
\caption{\textcolor{JungleGreen}{\textbf{Generative}} vs. \textcolor{darkgray}{\textbf{discriminative}} shape completion on the \emph{Automatica/YCB} dataset.}
\label{tab:ycb_cond_test}
\end{table}

\label{sec:ablations}
\noindent\textbf{Ablations.} To justify and inform our design choices, we perform an extensive ablation study on model size (Tab.~\ref{tab:model_size}), number of diffusion steps (Tab.~\ref{tab:diff_steps}), type of conditioning information (Tab.~\ref{tab:cond_type}), and the conditioning approach (Tab.~\ref{tab:bos_vs_cross}). We also provide an ablation on the number of completions accompanying Tab.~\ref{tab:kinect_cond_test}.

A single completion achieves results comparable to the discriminative model, while as few as two completions already outperform it. Including the results for ten completions from Tab.~\ref{tab:kinect_cond_test}, there is some indication of diminishing returns for larger numbers.

\begin{table}[htbp]
\centering
\begingroup
\sisetup{round-precision=2,table-format=2.2}
\begin{tabular}{lSSSB}
  \toprule
  & {$N=1$} & {$N=2$} & {$N=3$} & {$N=5$} \\
  \midrule
  Chamfer $\downarrow$ & 0.37772 & 0.36146 & 0.35474 & 0.34626 \\
  F1 $\uparrow$ & 41.0382 & 45.1204 & 47.0342 & 49.3725 \\
  Precision $\uparrow$ & 46.4972 & 51.0815 & 53.2610 & 55.9016 \\
  Recall $\uparrow$ & 38.7523 & 42.4165 & 44.104 & 46.1434 \\
  \bottomrule
\end{tabular}
\endgroup
\caption{Ablation on the number of generative completions $N$.}
\label{tab:num_completions}
\end{table}

To obtain the small models with approximately one-third of the parameters of the large variants, we simply halve the number of layers and the input dimension of all GeGLU activations. We find that the size of the model has a strong influence on latent generative modeling but not on auto-encoding (Tab.~\ref{tab:model_size}).

\begin{table}[htbp]
\centering
\begingroup
\sisetup{round-precision=2,table-format=2.2}
\begin{tabular}{lSS|lSS}
  \toprule
  & \multicolumn{2}{c|}{VAE} & {} & \multicolumn{2}{c}{Diffusion} \\
  \midrule
  & {Small} & {Large} & {} & {Small} & {Large} \\
  \midrule
  CD$\downarrow$ & \bfseries 0.0911306 & \bfseries 0.08685 & FID$\downarrow$ & 39.46212126426275 & \bfseries 32.61765585013054 \\
  F1$\uparrow$ & \bfseries 77.187 & 76.5176 & KID$\downarrow$ & 17.482928438063063 & \bfseries 12.998296084959962 \\
  Prec.$\uparrow$ & 82.6042 & \bfseries 83.3589 & Prec.$\uparrow$ & 41.72422680412371 & \bfseries 50.27036082474227 \\
  Rec.$\uparrow$ & \bfseries 74.528 & 72.5989 & Rec.$\uparrow$ & \bfseries 48.637500000000006 & 48.079123711340205 \\
  \bottomrule
\end{tabular}
\endgroup
\caption{Ablation on model size.}
\label{tab:model_size}
\end{table}

Recent diffusion models~\cite{karras2022elucidating} require only a fraction of the number of denoising steps during inference as the original DDPMs~\cite{ho2020denoising}. We follow~\citet{zhang20233dshape2vecset} and use as little as $18$ steps during inference. Nonetheless, we ablate this choice and find that doubling the number has a discernible impact while further increases show diminishing returns (Tab.~\ref{tab:diff_steps}).

\begin{table}[htbp]
\centering
\begingroup
\sisetup{round-precision=2,table-format=2.2}
\begin{tabular}{lSSSS}
  \toprule
  & {$T=18$} & {$T=35$} & {$T=50$} & {$T=100$} \\
  \midrule
  FID $\downarrow$ & 34.09504406121418 & 31.608463652792903 & 31.137713081573974 & \bfseries 30.92878054016137 \\
  KID $\times10^3\downarrow$ & 13.632151054429428 & 12.308178786161156 & 12.04857350700701 & \bfseries 11.888387090590589 \\
  Precision $\uparrow$ & 47.571134020618555 & 51.63634020618557 & 52.22551546391752 & \bfseries 52.70347938144331 \\
  Recall $\uparrow$ & 46.48621134020619 & \bfseries 46.706701030927833 & 46.63634020618558 & 46.52886597938144 \\
  \bottomrule
\end{tabular}
\endgroup
\caption{Ablation on the number of diffusion steps $T$.}
\label{tab:diff_steps}
\end{table}

We also investigate the impact of different feature types used for conditioning the generative models on the shape completion task. 
We find that the features from a model trained on classification are worse than those from a model trained for auto-encoding and, contrary to findings in~\citet{chen2020generative}, the features of the final layer are superior to those from the middle of the model for this task. Fine-tuning the feature extractor alongside the training of the generative model provides further improvement (Tab.~\ref{tab:cond_type}).

\begin{table}[htbp]
\setlength{\tabcolsep}{4pt}
\centering
\begin{tabular}{lS[round-precision=1,table-format=2.1]S[round-precision=1,table-format=2.1]|S[round-precision=1,table-format=3.1]S[round-precision=1,table-format=3.1]S[round-precision=1,table-format=3.1]}
  \toprule
  & {Class.} & {Recon.} & {Middle} & {Final} & {Final FT} \\
  \midrule
  FPD$\downarrow$ & 84.52282705044126 & \bfseries 70.26922976942728 & 129.1176543901172 & 112.9600401938194 & \bfseries 103.42393011717127 \\
  KPD$\downarrow$ & 7.3793381164318115 & \bfseries 5.892627758115871 & 12.223481492493294 & 10.385090639980401 & \bfseries 9.257547429556073 \\
  Prec.$\uparrow$ & 91.2109375 & \bfseries 92.27627840909091 & 87.39346590909091 & 88.40553977272727 & \bfseries 88.4765625 \\
  Rec.$\uparrow$ & 73.08238636363636 & \bfseries 77.61008522727273 & 60.91974431818182 & 68.07528409090909 & \bfseries 70.73863636363636 \\
  \bottomrule
\end{tabular}
\caption{Ablation on conditioning type: classification (class.) vs. reconstruction (recon.) and middle vs. final layer as well as final, fine-tuning (FT) features.}
\label{tab:cond_type}
\end{table}

Finally, while not competitive against the diffusion models on the shape completion from Kinect depth task, the beginning-of-sequence conditioning where we prepend the conditioning features to the latent code of the VQ-VAE consistently outperforms conditioning via cross-attention (Tab.~\ref{tab:bos_vs_cross}). This has the advantage that no additional cross-attention components must be added to the model, but it doubles the sequence length.

\begin{table}[htbp]
\centering
\begingroup
\sisetup{round-precision=2,table-format=3.2}
\begin{tabular}{lSS}
  \toprule
  & {BOS} & {Cross-Attn} \\
  \midrule
  1-NNA $\downarrow$ & \bfseries 55.26455965909091 & 55.75284090909091 \\
  FPD $\downarrow$ & \bfseries 167.57128272532714 & 181.96115229455313 \\
  KPD $\downarrow$ & \bfseries 17.571740634112864 & 19.41764932263356 \\
  Precision $\uparrow$ & 84.62357954545454 & \bfseries 85.7421875 \\
  Recall $\uparrow$ & \bfseries 43.66122159090909 & 39.683948863636365 \\
  \midrule
  Chamfer $\downarrow$ & \bfseries 0.426912 & 0.429941 \\
  F1 $\uparrow$ & \bfseries 36.4656 & 34.6072 \\
  Precision $\uparrow$ & \bfseries 39.1823 & 36.6419 \\
  Recall $\uparrow$ & \bfseries 35.6187 & 34.3232 \\
  \bottomrule
\end{tabular}
\endgroup
\caption{Ablation on Beginning-of-sequence (BOS) vs. cross-attention (Cross-Attn) conditioning during autoregressive training.}
\label{tab:bos_vs_cross}
\end{table}

\section{Conclusion}%
\label{sec:conclusion}

This work highlights the potential of generative modeling as an effective approach for high-fidelity 3D shape completion from single-view depth images. Through rigorous quantitative comparison to a discriminative method, we establish the advantage of generative models in effectively handling partial, noisy, and ambiguous input data for shape completion under realistic conditions, both regarding coverage of plausible alternatives and also accuracy in relation to a single ground truth complete shape. While this particular strength of generative models can be partially explained by their basic design, our empirical analysis uncovers details of their performance characteristics and also highlights key differences between latent diffusion-based and autoregressive approaches.

Limitations include the need to generate, and automatically select from, multiple completions to achieve optimal performance and a focus on specific model architectures, which may limit generalizability.

Future work will explore possible improvements in generative conditioning techniques such as Classifier-Free Guidance~\cite{ho2022classifier} and in quantized feature extraction from Residual VQ-VAEs~\cite{zeghidour2021soundstream} to unlock the full potential of autoregressive models in this domain.

{
    \small
    \bibliographystyle{ieeenat_fullname}
    \bibliography{main}
}
\appendix
\section{Implementation Details}
\label{sup:implementation_details}
Here, we extend Sec.~\ref{sec:method} from the main text to provide further details on the implementation and training. We train our VAEs for $800$ epochs with an effective batch size of $512$ and a learning rate of $4\times10^{-4}$ on four NVIDIA A100 80GB GPUs in less than a day; a fourth of the compute budget reported by~\citet{zhang20233dshape2vecset}. This is made possible through the reduction in model size (from $\sim$$106$ million to $\sim$$35$ million parameters), utilization of flash-attention~\cite{dao2022flashattention,dao2023flashattention} (native to $\mathrm{PyTorch}\geq2.2$), fused CUDA-kernels for NeRF encoding~\cite{tiny-cuda-nn}, GPU-accelerated farthest-point-sampling\footnote{\url{https://github.com/mit-han-lab/pvcnn/tree/master/modules}\label{fn:fps}} (FPS) and \texttt{bfloat16} mixed-precision training.

All latent generative models--both diffusion and autoregressive--have approx. the same size as the one in~\cite{zhang20233dshape2vecset} ($109$-$164$ million parameters) and are trained for $2000$ epochs with an effective batch size of $256$ and a learning rate of $10^{-4}$ on four A100 GPUs in less than two days; which again represents a fourth of the compute used by~\cite{zhang20233dshape2vecset}. We visualized the training progress, measured FID every $25$ epochs, and observed the majority of improvement occurring within the first $500$ epochs.

We find that while the VAEs are more sensitive to the \emph{range} of representable values, thus requiring \texttt{bfloat16} precision, the diffusion models require higher \emph{resolution} and, therefore, must be trained in \texttt{float16} precision to prevent divergence.

\section{Metrics}
\label{sup:metrics}
As discussed in the main text (Sec.~\ref{sub:metrics}), there is no clear consensus on the choice of evaluation metrics for 3D generative models, resulting in a great variety of metrics used. Additionally, their exact definitions and implementations can vary significantly. For this reason, this section provides the exact definition (or a reference to it) and additional details and discussion for all metrics used in our experiments.

\subsection{Instance-level}
\label{sup:instance_level}
These metrics rely on the comparison of individual instances, i.e., there is a one-to-one correspondence between prediction and ground truth, s.a. partial input and (best) completion.

\noindent\textbf{Volumetric Intersection-over-Union.}
\label{sup:iou}
The well-known \emph{Intersection-over-Union} metric, while ubiquitously used as a bounding-box measure in object detection, can also be defined for 3D volumes to evaluate implicit functions. We follow~\citet{mescheder2019occupancy} and compute the volumetric IoU for $10^5$ query points randomly sampled in a unit cube with additional total padding of $0.1$. It is restricted to watertight meshes and insensitive to fine details, especially at values below $50\%$~\cite{tatarchenko2019single} as well as oversensitive in low-volume regimes such as thin structures and walls~\cite{humt2023shape}. As a result, we primarily rely on other metrics for instance-level 3D shape comparisons.

\noindent\textbf{Chamfer Distance.}
\label{sup:chamfer_distance}
The (bidirectional, L2 or squared) Chamfer distance (CD) between two sets of points $\mathcal X$ and $\mathcal Y$  was introduced by ~\citet{fan2017point} and used compute COV and MMD~\cite{achlioptas2018learning} as well as 1-NNA~\cite{yang2019pointflow} as,
\begin{equation}
  \label{eq:chamfer_distance}
  \begin{aligned}
    \mathrm{Chamfer_{L2}}(\mathcal X,\mathcal Y) &= \sum_{x \in\mathcal X} \min_{y \in\mathcal Y} \|x - y\|_2^2\\ &+ \sum_{y \in\mathcal Y} \min_{x \in\mathcal X} \|x - y\|_2^2,
  \end{aligned}
\end{equation}
and later extended to an L1 variant in~\citet{mescheder2019occupancy} as the mean of an \emph{accuracy} and \emph{completeness} term,
\begin{equation}
  \label{eq:chamfer_distance_l1}
  \begin{aligned}
    \mathrm{Chamfer_{L1}}(\mathcal X,\mathcal Y) &= \frac{1}{2|\mathcal X|} \sum_{x \in\mathcal X} \min_{y \in\mathcal Y} \|x - y\|\\ &+ \frac{1}{2|\mathcal Y|} \sum_{y \in\mathcal Y} \min_{x \in\mathcal X} \|x - y\|
  \end{aligned}
\end{equation}
and, as in \cite{fan2017point,mescheder2019occupancy}, multiplied by \emph{``1/10 times the maximal edge length of the current object’s bounding box"} resulting in a factor of $10$.

We employ the L2 variant (eq.~\ref{eq:chamfer_distance}) when used within other metrics s.a. COV, MMD and 1-NNA--following their original definitions~\cite{achlioptas2018learning,yang2019pointflow}--but with $|\mathcal X|=|\mathcal Y|=2048$ farthest-point-samples to increase sensitivity and reduce variance--and the L1 (eq.~\ref{eq:chamfer_distance_l1}) variant with $|\mathcal X|=|\mathcal Y|=10^5$ random samples otherwise.
We found $2048$ FPS points to approximately resolve details of $10^4$ random points while significantly reducing computation time. We use GPU-accelerated implementations of both CD\footnote{\url{https://github.com/ThibaultGROUEIX/ChamferDistancePytorch}} and FPS. All point clouds for evaluation are sampled from the surface of generated and reference meshes.

\noindent\textbf{Earth Mover's Distance.}
\label{sup:emd}
While frequently recognized as a more precise alternative to CD, existing Earth Mover's Distance (EMD)~\cite{fan2017point} implementations almost exclusively rely on approximate solutions and thus do not guarantee correctness\footnote{\url{https://github.com/facebookresearch/pytorch3d/issues/211}}, and are still prohibitively slow for large-scale evaluations, even in their GPU-accelerated form\footnote{\url{https://github.com/Colin97/MSN-Point-Cloud-Completion/tree/master/emd}}. We, therefore, decide to omit EMD from our evaluation.

\noindent\textbf{F-score, Precision \& Recall.}
\label{sup:f_score}
First defined as a measure for multi-view 3D reconstruction quality~\cite{knapitsch2017tanks} and later introduced to 3D shape completion by~\citet{tatarchenko2019single}, the \emph{F-score} is the harmonic mean of \emph{precision} and \emph{recall}, where precision is the ratio of points in the completion that are close to the ground truth and recall is the ratio of points in the ground truth that are close to the completion. We use the default distance threshold of $0.01$ and $10^5$ surface samples for all evaluations.

\subsection{Set-level}
\label{sup:set_level}
These metrics compare two sets of instances, such as unconditional or class-conditional generations, against the train or test split or multiple completions to a single partial input. As explained in the previous section, all set-level metrics are computed on $2048$ FPS points.

\noindent\textbf{Coverage \& Minimum Matching Distance.}
\label{sup:cov_mmd}
For both \emph{Coverage} (COV) and \emph{Minimum Matching Distance} (MMD)~\cite{achlioptas2018learning}, we use the definition exactly as presented in~\citet{yang2019pointflow}. While neither their definition of CD nor MMD divide by the number of points, their code reveals\footnote{\url{https://github.com/stevenygd/PointFlow/blob/master/metrics/evaluation_metrics.py}}, that this average is indeed taken. In doing so, the influence of the number of points on the metrics is removed. We implement a batched, GPU-accelerated version for efficient paired-distance computation between all point clouds from two sets.

\noindent\textbf{Leave-One-Out 1-Nearest-Neighbor Accuracy.}
\label{sup:one_nna}
As for COV and MMD, we use the \emph{Leave-One-Out} (LOO) \emph{1-Nearest-Neighbor Accuracy} 1-NNA definition of~\citet{yang2019pointflow} who proposed it as a more reliable alternative to the former. While for unconditional and class-conditional generative models, a score of $50\%$ denotes peak performance, we point out that for instance-conditioned tasks, such as shape completion, a perfect model would achieve $0\%$, as the LOO NN to the ground truth shape should always be the generated completion.

\noindent\textbf{Edge Count Difference.}
\label{sup:ecd}
We use the definition and implementation\footnote{\url{https://github.com/GregorKobsik/Octree-Transformer/blob/master/evaluation/evaluation.py}} by~\citet{ibing20213d}, who also recognized the shortcomings of COV and MMD and propose \emph{Edge Count Difference} (ECD) as another alternative. We found that ECD frequently yields contrary results to all other metrics, thus making it seem less reliable than, e.g., 1-NNA.

\noindent\textbf{Total Mutual Difference.}
\label{sup:tmd}
Designed as a \emph{diversity} measure by~\citet{wu2020multimodal}, the \emph{Total Mutual Difference} (TMD) for a partial input is the sum of the LOO CD between $10$ completions.

\noindent\textbf{Unidirectional Hausdorff Distance.}
\label{sup:uhd}
The \emph{Unidirectional Hausdorff Distance} (UHD)~\cite{wu2020multimodal}, on the other hand, is supposed to measure \emph{fidelity} as the average distance from $10$ completions to the partial input.

\noindent\textbf{Fréchet \& Kernel Pointcloud Distance.}
\label{sup:fpd_kpd}
Instead of in metric space, one can also compare point clouds in the higher-dimensional feature space of a pre-trained neural network to potentially capture high-level semantic information. To this end, \citet{shue20233d} define a derivative of the Fréchet Inception Distance (FID)~\cite{heusel2017gans} as the \emph{Fréchet Pointcloud Distance} (FPD) between two sets of point clouds. Similarly, \citet{zhang20233dshape2vecset} propose \emph{Kernel Pointcloud Distance} as a derivative of the \emph{Kernel Inception Distance} (KID)~\cite{binkowski2018demystifying}. We use the same $2048$ FPS points to compute FPD and KPD as used for all other set-level metrics and our pre-trained VAE to extract point features. We reuse low-level functionality from the \texttt{clean-fid}~\cite{parmar2021cleanfid} Python package.

\noindent\textbf{Fréchet \& Kernel Inception Distance.}
\label{sup:fid_kid}
The \emph{Fréchet Inception Distance}~\cite{heusel2017gans} computes the Fréchet distance between two Gaussian distributions in the feature space of the \emph{Inception V3}~\cite{szegedy2015rethinking} network pre-trained on the \emph{ImageNet}~\cite{deng2009imagenet} dataset. Therefore, two implicit assumptions are made: (1) The feature space follows a Gaussian distribution, and (2) the images ingested by the Inception V3 network are identically distributed to the ImageNet dataset. The more these assumptions are violated, the less reliable FID becomes~\cite{kynkaanniemi2022role}.

The second assumption can be somewhat alleviated through the use of a different pre-trained network, potentially trained on a larger and more diverse dataset such as CLIP~\cite{radford2021learning} features from a Vision Transformer~\cite{dosovitskiy2020image} as proposed in~\citet{kynkaanniemi2022role}. We refer to this metric as $\mathrm{FID}_{\mathrm{CLIP}}$.

The \emph{Kernel Inception Distance}~\cite{binkowski2018demystifying} is a non-parametric alternative to FID, which uses the \emph{Maximum Mean Discrepancy}~\cite{gretton2012kernel} to compare the feature distributions of two sets of images and therefore relaxes the Gaussian assumption.

To measure the perceptual quality of 3D data, FID and KID are adapted to the 3D domain by~\citet{zheng2022sdf} and \citet{zhang20233dshape2vecset} respectively through rendering of shaded images from $20$ uniformly distributed viewpoints around the object. \emph{Shading-image-based} FID and KID are the average FID and KID across all views.

\noindent\textbf{FID decompositions.}
\label{sup:fid_decomp}
Finally, \citet{sajjadi2018assessing} propose a decomposition of FID into \emph{Precision} and \emph{Recall}, improved upon by~\citet{kynknniemi2019improved}, which is the definition we use throughout this work.

\citet{ferjad2020reliable} acknowledge the improvements made by~\citet{kynknniemi2019improved} but find remaining failure cases of the improved precision and recall formulations and therefore propose \emph{Density} and \emph{Coverage} as drop-in replacements.

We further propose to also decompose FPD to obtain an even more detailed view of the generative performance of 3D data.

\subsection{Recommendations}
\label{sup:recommendations}
Based on our extensive empirical evaluation and literature review, we recommend the following metrics for the evaluation of 3D generative models in general and the shape completion task in particular:

\begin{itemize}
  \item For \emph{instance-level} evaluation, we only recommend the F1-score but highly recommend the precision and recall decomposition. All other metrics in this category, like CD, EMD, and IoU, feature at least one highly problematic aspect, as discussed in their dedicated sections.
  \item For \emph{set-level} evaluation, we strongly recommend KPD and FPD, especially with a task-specific feature extractor (ideally a VAE), but shading-image-based FID and KID are viable alternatives. For both FID and FPD, we recommend the (improved) precision and recall decomposition to gain valuable insights into the origin of the observed performance. The only non-feature-based metric we recommend is 1-NNA.
\end{itemize}

\section{Additional Results}
\label{sup:additional_results}

\subsection{Quantitative Results}
\label{sup:quantitative_results}

\begin{table}[htbp]
\centering
\sisetup{
  round-precision=3,
  table-format=2.3
}
\begin{tabular}{lSS}
  \toprule
  & {Normal Consistency~\cite{mescheder2019occupancy} $\uparrow$} & {IoU $\uparrow$} \\
  \midrule
  VAE & \bfseries 95.9662 & \bfseries 93.6347 \\
  VQ-VAE & 92.0651 & 85.4528 \\
  \bottomrule
\end{tabular}
\caption{Reconstruction quality; class average. Watertight meshes only. Extends Tab.~\ref{tab:recon_quality}.}
\label{tab:recon_quality_sup}
\end{table}

\begin{table}[htbp]
\setlength{\tabcolsep}{4pt}
\centering
\sisetup{
  round-precision=1,
  table-format=2.1
}
\begin{tabular}{lSSSSSSHHH}
  \toprule
  & {D=32} & {kmeans} & {N=16k} & {revive} & {sample} & {IoU $\uparrow$} & {F1 $\uparrow$} & {Prec. $\uparrow$} & {Rec. $\uparrow$} \\
  \midrule
  FSQ & {} & {} & {} & {} & {} & 81.18 & 89.51 & 89.57 & 89.48 \\
  LFQ & {} & {} & {} & {} & {} & 79.93 & 88.73 & 89.15 & 88.34 \\
  \midrule
  VQ & {} & {} & {} & {} & {} & 78.91 & 88.09 & 88.09 & 88.12 \\
     & {$\checkmark$} & {} & {} & {} & {} & 81.27 & 89.57 & 89.87 & 89.30 \\
     & {} & {$\checkmark$} & {} & {} & {} & 85.92 & 92.38 & 92.45 & 92.32 \\
     & {} & {$\checkmark$} & {$\checkmark$} & {} & {} & 89.23 & 94.28 & 94.26 & 94.31 \\
     & {} & {$\checkmark$} & {$\checkmark$} & {$\checkmark$} & {} & 88.76 & 93.99 & 94.28 & 93.70 \\
     & {} & {$\checkmark$} & {$\checkmark$} & {} & {$\checkmark$} & 89.33 & 94.34 & 94.67 & 94.03 \\
  \bottomrule
\end{tabular}
\caption{VQ-VAE ablations.}
\label{tab:vqvae_ablations}
\end{table}

\begin{table}[htbp]
\centering
\scriptsize
\setlength{\tabcolsep}{2pt}
\sisetup{
  round-precision=2,
  table-format=4.2
}
\resizebox{\linewidth}{!}{%
\begin{tabular}{llSS[round-precision=3,table-format=1.3]HS[round-precision=0,table-format=4.0]HHHHSS}
  \toprule
  & {} & {COV$\uparrow$} & {MMD$\downarrow$} & {1-NNA$\downarrow$} & {ECD $\downarrow$} & {FPD$\downarrow$} & {KPD$\downarrow$} & {Prec.$\uparrow$} & {Rec.$\uparrow$} & {Dens.$\uparrow$} & {Cov.$\uparrow$} \\
  \midrule
  \multirow{3}{*}{\rotatebox{90}{\textbf{Chair}}}
  & LAS-Diff. & 45.790251107828656 & \bfseries 3.5219704658357065 & 59.08419497784343 & 80.49428184979755 & 99.17017632099714 & 9.312835463831902 & \bfseries 96.89807976366323 & 63.36779911373708 & \bfseries 1.3031019202363368 & 0.8892171344165436 \\
  & 3DS2VS & \bfseries 51.55096011816839 & 3.530962519678514 & 58.93648449039882 & 25.97728184206988 & 94.01398448338341 & 7.1597585662331475 & 85.67208271787297 & \bfseries 77.10487444608567 & 0.7878877400295421 & 0.8286558345642541 \\
  & Ours & 50.81240768094535 & 3.5883812063567592 & \bfseries 58.49335302806499 & \bfseries 7.057903294490758 & \bfseries 89.58658797523253 & \bfseries 6.97410854214546 & 95.56868537666174 & 60.70901033973413 & 1.3028064992614476 & \bfseries 0.9069423929098966 \\
  \midrule
  \multirow{3}{*}{\rotatebox{90}{\textbf{Plane}}}
  & LAS-Diff. & 38.11881188118812 & 1.2491523706521907 & 82.67326732673267 & 164.4906044974176 & 257.6618045508026 & 34.79479911264624 & 75.0 & 11.386138613861387 & 0.43514851485148515 & 0.3217821782178218 \\
  & 3DS2VS & 48.267326732673266 & 1.0588670675763814 & 69.67821782178217 & 37.3497194645217 & 165.00836229770084 & 22.3454074706489 & 68.31683168316832 & \bfseries 34.15841584158416 & 0.4618811881188119 & 0.4628712871287129 \\
  & Ours & \bfseries 50.000 & \bfseries 1.057608307098352 & \bfseries 69.05940594059405 & \bfseries 10.22629097498548 & \bfseries 139.68486469301092 & \bfseries 17.054163521795438 & \bfseries 84.15841584158416 & 30.94059405940594 & \bfseries 0.7217821782178219 & \bfseries 0.556930693069307 \\
  \midrule
  \multirow{3}{*}{\rotatebox{90}{\textbf{Car}}}
  & LAS-Diff. & 28.57142857142857 & \bfseries 0.9919305105702341 & 86.3150867823765 & \bfseries 482.99944047975106 & 99.01572031076898 & \bfseries 16.26722705225314 & \bfseries 62.61682242990654 & \bfseries 55.67423230974633 & 0.27476635514018694 & \bfseries 0.3591455273698264\\
  & 3DS2VS & 25.500667556742324 & 1.230789715447805 & 91.0547396528705 & 2035.9806491919655 & 170.99114959950953 & 27.70528120088605 & 60.88117489986649 & 39.65287049399199 & 0.22429906542056074 & 0.17623497997329773 \\
  & Ours & \bfseries 37.383177570093457 & 1.0882384674632934 & \bfseries 82.1762349799733 & 538.104533054052 & \bfseries 84.73930947206367 & 16.61974036928133 & 59.41255006675568 & 48.19759679572764 & \bfseries 0.27716955941255006 & 0.3030707610146863 \\
  \midrule
  \multirow{3}{*}{\rotatebox{90}{\textbf{Table}}}
  & LAS-Diff. & 49.88235294117647 & \bfseries 3.1105742655927315 & 55.35294117647059 & 135.57997966311905 & 158.87359836247379 & 19.95190206950745 & 94.58823529411765 & \bfseries 72.11764705882353 & 0.9985882352941178 & 0.8635294117647059 \\
  & 3DS2VS & 50.94117647058823 & 3.2490435972303043 & 56.76470588235294 & 19.636860658267743 &  148.09748139253043& 15.082496804238199 & 92.47058823529412 & 71.52941176470589 & 0.9272941176470589 & 0.8294117647058824 \\
  & Ours & \bfseries 52.8235294117647 & 3.1874982838053256 & \bfseries 53.70588235294118 & \bfseries 15.702286851771053 & \bfseries 128.35344985256415 & \bfseries 9.533235844308027 & \bfseries 96.70588235294117 & 67.64705882352942 & \bfseries 1.203764705882353 & \bfseries 0.8717647058823529 \\
  \midrule
  \multirow{3}{*}{\rotatebox{90}{\textbf{Rifle}}}
  & LAS-Diff. & 32.489451476793246 & 0.9503240209504597 & 77.42616033755274 & 180.3133876986132 & 693.8384515518387 & 115.62660763954709 & \bfseries 96.62447257383966 & 34.59915611814346 & 0.5341772151898735 & 0.21518987341772153 \\
  & 3DS2VS & 45.147679324894513 & \bfseries 0.8467293172148312 & \bfseries 66.03375527426161 & 38.638367919480075 & 418.0064691086095 & 57.748641784765674 & 91.9831223628692 & \bfseries 50.63291139240507 & 0.7147679324894516 & 0.4219409282700422 \\
  & Ours & \bfseries 46.41350210970464 & 0.8946201745749801 & 70.46413502109705 & \bfseries 11.755426656071002 & \bfseries 347.778028072869 & \bfseries 52.889068495706994 & 95.78059071729957 & 30.80168776371308 & \bfseries 0.8067510548523207 & \bfseries 0.4472573839662447 \\
  \toprule
  \multirow{3}{*}{\rotatebox{90}{\textbf{Mean}}}
  & LAS-Diff. & 38.97045919568301 & 1.9647903267202644 & 72.17033012019519 & 208.7755388377397 & 261.71195021937626 & 39.190674267557156 & 85.14552201230542 & 47.428994642862354 &  0.7091564481424 & 0.5297728250373238 \\
  & 3DS2VS & 44.281562040613345 & 1.9832784434295672 & 68.49358062433322 & 431.516575815261 & 199.22348937634675 & 26.00831716535439 & 79.86475997981421 & \bfseries 54.61569678775455 & 0.623226008741085 & 0.5438229589284378 \\
  & Ours & \bfseries 47.48652335450163 & \bfseries 1.9632692878597422 & \bfseries 66.77980226453412 & \bfseries 116.56928816627405 & \bfseries 158.02844801314805 & \bfseries 20.61406335464745 & \bfseries 86.32522487104847 & 47.659189556422044 & \bfseries 0.8624547995252986 & \bfseries 0.6171931873684975 \\
  \bottomrule
\end{tabular}%
}
\caption{Comparison of \emph{class-conditional} generative models. MMD$\times10^3$. Extends Tab.~\ref{tab:class_cond_test}.}
\label{tab:class_cond_test_sup}
\end{table}

\begin{table}[htbp]
\centering
\sisetup{
  round-precision=3,
  table-format=3.3
}
\begin{tabular}{lSS}
  \toprule
  & {\normalfont Diffusion (VAE)} & {AR (VQ-VAE)} \\
  \midrule
  $\mathrm{FID}_{\mathrm{CLIP}} \downarrow$ & 3.5967370423096767 & \bfseries 3.5806686852101683 \\
  Density $\uparrow$ & \bfseries 0.30329819587628865 & 0.2925711340206185 \\
  Coverage $\uparrow$ & \bfseries 0.3300167525773196 & 0.2915154639175258 \\
  \midrule
  1-NNA $\uparrow$ & \bfseries 63.93821022727273 & 68.13742897727273 \\
  FPD $\downarrow$ & \bfseries 74.42017447622948 & 79.42524573050878 \\
  KPD $\downarrow$ & \bfseries 4.19842792086049 & 4.918914700430652 \\
  Precision $\uparrow$ & \bfseries 56.55847738139022 & 56.04450165502024 \\
  Recall $\uparrow$ & \bfseries 59.65336520779699 & 54.3940787054064 \\
  COV $\uparrow$ & \bfseries 48.33096590909091 & 45.41903409090909 \\
  MMD$\times10^3\downarrow$ & 2.382356397849218 & \bfseries 2.3443228416187944 \\
  ECD $\downarrow$ & \bfseries 60.01978722529639 & 124.5683891421007 \\
  Density $\uparrow$ & \bfseries 1.0263139204545455 & 1.0133522727272728 \\
  Coverage $\uparrow$ & \bfseries 0.7487571022727273 & 0.7230113636363636 \\
  \bottomrule
\end{tabular}
\caption{Comparison of diffusion and autoregressive \emph{unconditional} generative shape modeling on continuous (VAE) and discrete (VQ-VAE) latents. Extends Tab.~\ref{tab:uncond_gen}.}
\label{tab:uncond_gen_sup}
\end{table}

\begin{table}[htbp]
\centering
\sisetup{
  round-precision=3,
  table-format=3.3
}
\setlength{\tabcolsep}{4pt}
\begin{tabular}{lSS|S}
  \toprule
  & \multicolumn{2}{c}{{VQ-VAE}} & {VAE} \\
  \midrule
  & {Diffusion} & {Autoregressive} & {Diffusion} \\
  \midrule
  $\mathrm{FID}_{\mathrm{CLIP}} \downarrow$ & 4.675047093827677 & \bfseries 3.318907404164384 & 3.153972887895471 \\
  Density $\uparrow$ & 0.18877731958762886 & \bfseries 0.3011889175257732 & 0.3382208762886598 \\
  Coverage $\uparrow$ & 0.1953273195876289 & \bfseries 0.3058981958762887 & 0.35038530927835054 \\
  \midrule
  COV $\uparrow$ & 46.129261363636365 & \bfseries 47.15909090909091 & 48.277698863636365 \\
  MMD$\times10^3\downarrow$ & 2.458678500918733 & \bfseries 2.314231863834948 & 2.3485657634681315 \\
  ECD $\downarrow$ & 127.99980766351999 & \bfseries 102.31735937176794 & 73.03440617352317 \\
  Density $\uparrow$ & \bfseries 1.0762428977272727 & 0.9767400568181819 & 1.0092329545454546 \\
  Coverage $\uparrow$ & \bfseries 0.7455610795454546 & 0.7210582386363636 & 0.7464488636363636 \\
  \bottomrule
\end{tabular}
\caption{Comparison of diffusion and autoregressive \emph{class-conditional} generative shape modeling on the same latent space. Extends Tab.~\ref{tab:class_cond_gen}.}
\label{tab:class_cond_gen_sup}
\end{table}

\begin{table*}[htbp]
  \centering
  \scriptsize
  \setlength{\tabcolsep}{2.5pt}
  \sisetup{
    round-precision=2,
    table-format=4.2
  }
  \resizebox{\textwidth}{!}{%
  \begin{tabular}{lHHSSS[round-precision=3,table-format=1.3]S[round-precision=3,table-format=1.3]S[round-precision=0,table-format=4.0]S[round-precision=0,table-format=4.0]HHHHSSSSHHHHHHHHS[round-precision=3,table-format=1.3]S[round-precision=3,table-format=1.3]}
    \toprule
    & \multicolumn{2}{H}{{1-NNA $\downarrow$}} & \multicolumn{2}{c}{{COV $\uparrow$}} & \multicolumn{2}{c}{{MMD$\times10^3\downarrow$}} & \multicolumn{2}{c}{{ECD $\downarrow$}} & \multicolumn{2}{H}{{FPD $\downarrow$}} & \multicolumn{2}{H}{{KPD $\downarrow$}} & \multicolumn{2}{c}{{Prec. $\uparrow$}} & \multicolumn{2}{c}{{Rec. $\uparrow$}} & \multicolumn{2}{H}{{CD $\downarrow$}} & \multicolumn{2}{H}{{F1 $\uparrow$}} & \multicolumn{2}{H}{{Prec. $\uparrow$}} & \multicolumn{2}{H}{{Rec. $\uparrow$}} & {TMD $\uparrow$} & {UHD $\downarrow$} \\
    \midrule
    & {\color{darkgray}\textbf{D}} & {\color{JungleGreen}\textbf{G}} & {\color{darkgray}\textbf{D}} & {\color{JungleGreen}\textbf{G}} & {\color{darkgray}\textbf{D}} & {\color{JungleGreen}\textbf{G}} & {\color{darkgray}\textbf{D}} & {\color{JungleGreen}\textbf{G}} & {\color{darkgray}\textbf{D}} & {\color{JungleGreen}\textbf{G}} & {\color{darkgray}\textbf{D}} & {\color{JungleGreen}\textbf{G}} & {\color{darkgray}\textbf{D}} & {\color{JungleGreen}\textbf{G}} & {\color{darkgray}\textbf{D}} & {\color{JungleGreen}\textbf{G}} & {\color{darkgray}\textbf{D}} & {\color{JungleGreen}\textbf{G}} & {\color{darkgray}\textbf{D}} & {\color{JungleGreen}\textbf{G}} \\
    Chair & 38.25701624815362 & \bfseries 33.677991137370755 & 62.62924667651403 & \bfseries 66.76514032496307 & 2.67122159783379 & \bfseries 2.4602482868912195 & 118.40910474143494 & \bfseries 95.36589650748068 & 299.77586420908483 & \bfseries 128.3521815246902 & 48.546258758311104 & \bfseries 16.26660255961808 & 80.35450516986706 & \bfseries 87.74002954209749 & 48.892171344165436 & \bfseries 85.81979320531757 & 0.395799 & \bfseries 0.32730 & 43.2726 & \bfseries 52.4714 & 47.1179 & \bfseries 56.6614 & 41.0942 & \bfseries 50.4970 & 3.685 & 6.5898 \\
    Plane & 60.27227722772277 & \bfseries 50.99009900990099 & 58.91089108910891 & \bfseries 60.64356435643564 & 0.9370196609176341 & \bfseries 0.8174630153493721 & 119.06117707569913 & \bfseries 72.73906505499616 & 425.6293969581641 & \bfseries 213.86719934466237 & 62.18351376616974 & \bfseries 28.971572504498745 & 75.74257425742574 & \bfseries 85.39603960396039 & 14.85148514851485 & \bfseries 48.267326732673266 & 0.311524 & \bfseries 0.29014 & 47.0918 & \bfseries 55.6980 & 50.6998 & \bfseries 62.4971 & 45.3234 & \bfseries 51.8630 & 2.3013 & 4.9385 \\
    Car & 88.18424566088118 & \bfseries 74.6995994659546 & 31.37516688918558 & \bfseries 44.85981308411215 & 1.2129643377574418 & \bfseries 1.032585668315718 & 1445.4602492654508 & \bfseries 394.88978063394995 & 172.11391390453423 & \bfseries 114.32870706500916 & 33.884359429709235 & \bfseries 17.852929726043726 & 27.369826435246997 & \bfseries 49.265687583444595 & 12.016021361815754 & \bfseries 54.47263017356475 & 0.282849 & \bfseries 0.25958 & 38.118 & \bfseries 47.6417 & 50.4374 & \bfseries 58.9671 & 31.1896 & \bfseries 40.9205 & 2.8457 & 5.5513 \\
    Table & 38.941176470588235 & \bfseries 35.94117647058824 & 62.11764705882353 & \bfseries 65.41176470588236 & 2.4370308750761016 & \bfseries 2.3344851503326724 & 35.82915101804676 & \bfseries 16.37833482094103 & 274.44469570192086 & \bfseries 134.06604916980314 & 31.130998642888883 & \bfseries 13.570355501565526 & 84.82352941176471 & \bfseries 95.76470588235294 & 64.11764705882353 & \bfseries 79.17647058823529 & 0.447081 & \bfseries 0.26435 & 48.0384 & \bfseries 57.1465 & 49.9848 & \bfseries 60.3705 & 48.2835 & \bfseries 56.1067 & 4.6604 & 5.57 \\
    Rifle & 57.38396624472574 & \bfseries 54.21940928270043 & 49.78902953586498 & \bfseries 53.16455696202531 & \bfseries 0.6971431429815215 & 0.6977838673088543 & 82.89059439114324 & \bfseries 73.63435002288752 & 572.242164205503 & \bfseries 479.7763144782542 & 85.24054806676735 & \bfseries 69.67679097630563 & 75.10548523206751 & \bfseries 91.13924050632911 & \bfseries 40.92827004219409 & 33.755274261603374 & 0.550823 & \bfseries 0.54209 & 37.3875 & \bfseries 45.9956 & 41.2176 & \bfseries 55.5294 & 35.8422 & \bfseries 41.5188 & 3.1319 & 4.9767 \\
    \midrule
    Mean & 56.607736 & \bfseries 49.905655073303 & 52.9643962498994 & \bfseries 58.16896788668371 & 1.5910759229132978 & \bfseries 1.4685131976395673 & 360.330055 & \bfseries 130.60148540805108 & 348.841207 & \bfseries 214.07809031648384 & 52.197136 & \bfseries 29.267650253606337 & 68.679184 & \bfseries 81.8611406236369 & 36.161119 & \bfseries 60.298298992278866 & 0.397615 & \bfseries 0.336692 & 42.78166 & \bfseries 51.79064 & 47.8915 & \bfseries 58.8051 & 40.34658 & \bfseries 48.18120 & 3.3248599999999997 & 5.525260000000001 \\
    \midrule
    All & 53.59552556818182 & \bfseries 48.52627840909091 & 56.35653409090909 & \bfseries 60.49360795454546 & 1.8729592749157189 & \bfseries 1.7846501545555136 & \bfseries 189.27667301350087 & 238.492222159629 & 203.88417518397864 & \bfseries 103.42393011717127 & 24.347399180570868 & \bfseries 9.257547429556073 & 77.16619318181818 & \bfseries 88.4765625 & 41.08664772727273 & \bfseries 70.73863636363636 & / & / & / & / & / & / & / & / & / & / \\
    \bottomrule
  \end{tabular}
  }
  \caption{{\color{JungleGreen}\textbf{Generative}} ({\color{JungleGreen}\textbf G}) vs. {\color{darkgray}\textbf{discriminative}} ({\color{darkgray}\textbf D}) shape completion from a single Kinect depth image. TMD and UHD from 10 generations. Extends Tab.~\ref{tab:kinect_cond_test}.}
  \label{tab:kinect_cond_test_sup}
\end{table*}

\subsection{Qualitative Results}
\label{sup:qualitative_results}

\begin{figure}[t]
    \centering
    \begin{subfigure}{0.25\linewidth}
        \includegraphics[width=\textwidth]{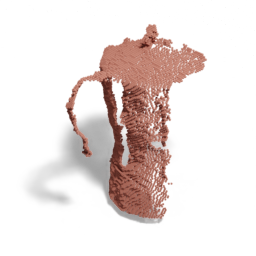}%
    \end{subfigure}%
    \begin{subfigure}{0.25\linewidth}
        \includegraphics[width=\textwidth]{figures/real/mokka/gt.png}%
    \end{subfigure}%
    \begin{subfigure}{0.25\linewidth}
        \includegraphics[width=\textwidth]{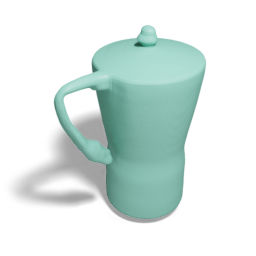}%
    \end{subfigure}%
    \begin{subfigure}{0.25\linewidth}
        \includegraphics[width=\textwidth]{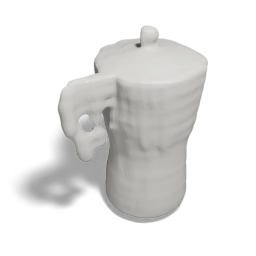}%
    \end{subfigure}%

    \begin{subfigure}{0.25\linewidth}
        \includegraphics[width=\textwidth]{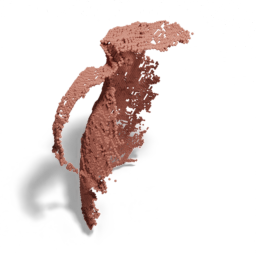}%
    \end{subfigure}%
    \begin{subfigure}{0.25\linewidth}
        \includegraphics[width=\textwidth]{figures/real/thermos/gt.png}%
    \end{subfigure}%
    \begin{subfigure}{0.25\linewidth}
        \includegraphics[width=\textwidth]{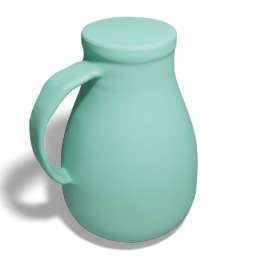}%
    \end{subfigure}%
    \begin{subfigure}{0.25\linewidth}
        \includegraphics[width=\textwidth]{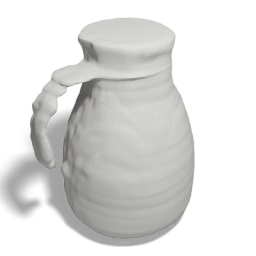}%
    \end{subfigure}%

    \begin{subfigure}{0.25\linewidth}
        \includegraphics[width=\textwidth]{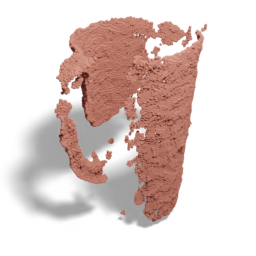}%
    \end{subfigure}%
    \begin{subfigure}{0.25\linewidth}
        \includegraphics[width=\textwidth]{figures/real/pitcher/gt.png}%
    \end{subfigure}%
    \begin{subfigure}{0.25\linewidth}
        \includegraphics[width=\textwidth]{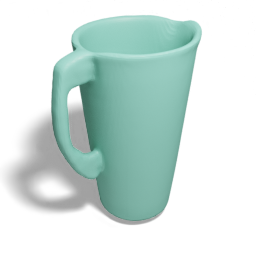}%
    \end{subfigure}%
    \begin{subfigure}{0.25\linewidth}
        \includegraphics[width=\textwidth]{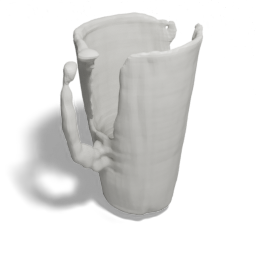}%
    \end{subfigure}%

    \begin{subfigure}{0.25\linewidth}
        \includegraphics[width=\textwidth]{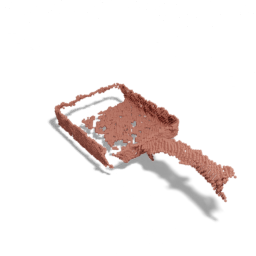}%
    \end{subfigure}%
    \begin{subfigure}{0.25\linewidth}
        \includegraphics[width=\textwidth]{figures/real/pan/gt.png}%
    \end{subfigure}%
    \begin{subfigure}{0.25\linewidth}
        \includegraphics[width=\textwidth]{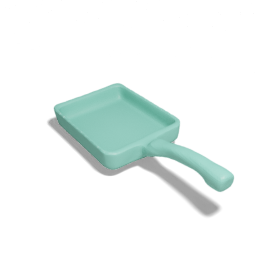}%
    \end{subfigure}%
    \begin{subfigure}{0.25\linewidth}
        \includegraphics[width=\textwidth]{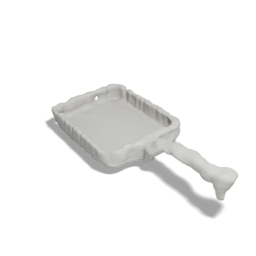}%
    \end{subfigure}%

    \begin{subfigure}{0.25\linewidth}
        \includegraphics[width=\textwidth]{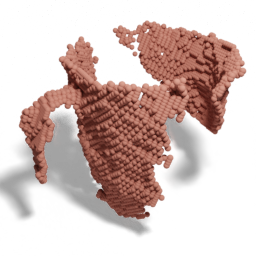}%
    \end{subfigure}%
    \begin{subfigure}{0.25\linewidth}
        \includegraphics[width=\textwidth]{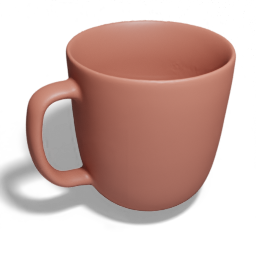}%
    \end{subfigure}%
    \begin{subfigure}{0.25\linewidth}
        \includegraphics[width=\textwidth]{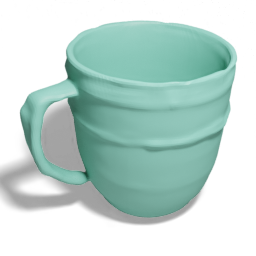}%
    \end{subfigure}%
    \begin{subfigure}{0.25\linewidth}
        \includegraphics[width=\textwidth]{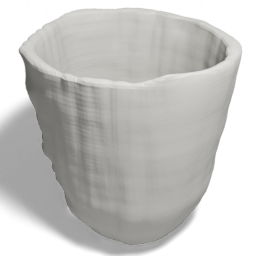}%
    \end{subfigure}%
    \caption{Real-world examples using depth data from a Kinect sensor. From left to right: {\color{BrickRed}\textbf{input}}, {\color{BrickRed}\textbf{ground truth}}, {\color{JungleGreen}\textbf{generative}} (best), and {\color{darkgray}\textbf{discriminative}}.}
    \label{fig:real_world}
\end{figure}

\end{document}